\pdfoutput=1

\documentclass[11pt]{article}

\usepackage{acl}

\usepackage{times}
\usepackage{pifont}
\usepackage{latexsym}
\usepackage{booktabs}
\usepackage{graphicx}
\usepackage{times}
\usepackage{latexsym}
\usepackage{caption}
\usepackage{subcaption}
\usepackage{colortbl}
\usepackage{courier}
\usepackage[T1]{fontenc}
\usepackage{amsmath}
\usepackage{amssymb}
\usepackage{multirow}
\usepackage{dsfont}
\usepackage{url}
\usepackage{makecell}
\usepackage{xcolor}
\usepackage{bm}
\usepackage{tikz}
\usepackage{url}
\usepackage{relsize}
\usepackage{hyperref}
\usepackage{enumitem}
\usepackage{makecell}
\usepackage{fontawesome}

\usepackage[T1]{fontenc}

\usepackage[utf8]{inputenc}

\usepackage{microtype}
\newcommand{\datasetname}{\textbf{\textsc{ComBO}}}
\newcommand{\xmark}{{\color{magenta} \ding{55}}}%

\title{\datasetname:  A Complete Benchmark for Open KG Canonicalization}

\makeatletter
\def\@fnsymbol#1{\ensuremath{\ifcase#1\or \dagger\or *\or \ddagger\or
   \mathsection\or \mathparagraph\or \|\or **\or \dagger\dagger
   \or \ddagger\ddagger \else\@ctrerr\fi}}
\makeatother

\newcommand{\shtu}{\textsuperscript{\faSunO}}
\newcommand{\shrcvi}{\textsuperscript{\faMoonO}}
\newcommand{\damo}{\textsuperscript{\faStarO}}

\newcommand{\code}{\url{https://github.com/jeffchy/COMBO/tree/main}}

\author{
    \textbf{Chengyue Jiang}\shtu \shrcvi \thanks{$~~$This work was done during Chengyue Jiang's internship at DAMO Academy, Alibaba Group. } ,
    \textbf{Yong Jiang}\damo$^{\ast}$,
    \textbf{Weiqi Wu}\shtu,
    \textbf{Yuting Zheng}\shtu, \\
    \textbf{Pengjun Xie}\damo,
    \textbf{Kewei Tu}\shtu \shrcvi \thanks{$~~$Yong Jiang and Kewei Tu are corresponding authors.} \\
    \shtu School of Information Science and Technology, ShanghaiTech University \\
    \shrcvi Shanghai Engineering Research Center of Intelligent Vision and Imaging \\
    \damo DAMO Academy, Alibaba Group, China \\
    \texttt{\{jiangchy,tukw,wuwq\}@shanghaitech.edu.cn;}\\
    \texttt{\{yongjiang.jy,chengchen.xpj\}@alibaba-inc.com }
}

\begin{document}
\maketitle
\begin{abstract}
Open knowledge graph (KG) consists of {\it (subject, relation, object)} triples extracted from millions of raw text. The {\it subject} and {\it object} noun phrases and the {\it relation} in open KG have severe redundancy and ambiguity and need to be canonicalized. 
Existing datasets for open KG canonicalization only provide gold entity-level canonicalization for noun phrases. In this paper, we present \datasetname , a \textbf{Com}plete \textbf{B}enchmark for \textbf{O}pen KG canonicalization. Compared with existing datasets, we additionally provide gold canonicalization for relation phrases, gold ontology-level canonicalization for noun phrases, as well as source sentences from which triples are extracted. We also propose metrics for evaluating each type of canonicalization. On the \datasetname\ dataset, we empirically compare previously proposed canonicalization methods as well as a few simple baseline methods based on pretrained language models. We find that properly encoding the phrases in a triple using pretrained language models results in better relation canonicalization and ontology-level canonicalization of the noun phrase. We release our dataset, baselines, and evaluation scripts at \code.
\end{abstract}
\section{Introduction}
\begin{figure}[h]
    \centering
    \scalebox{0.25}{
    \includegraphics{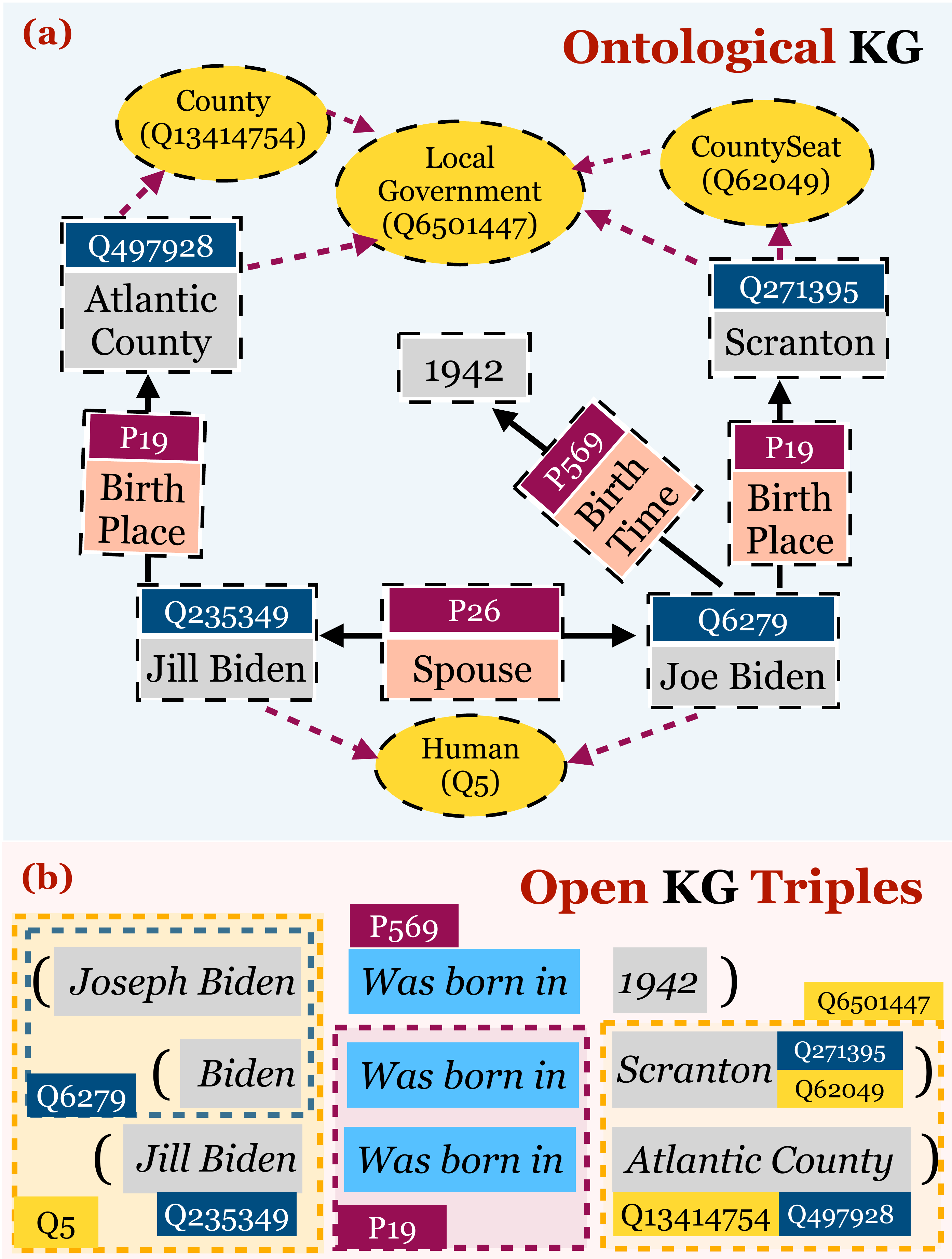}}
    \caption{Example of ontological KG (a) and Open KG triples (b). The differently colored bounding boxes and the tags on the open KG triples illustrate three types of gold canonicalization. Yellow (e.g., \emph{Q5} Human) shows the gold ontology-level NP cluster, the salvia blue (e.g., \emph{Q6279}) indicates the gold entity-level NP cluster, and the purple (e.g., P19) indicates gold RP cluster.}
    \label{fig:canonic}
\end{figure}
Large ontological knowledge graphs (KG) such as Wikidata \cite{Wikidata}, DBpedia \cite{DBpedia}, Freebase \cite{FreeBase} use a complex ontology to formalize and organize all the entities and relations. Figure \ref{fig:canonic}(a) shows an example ontological knowledge graph (Wikidata): ``\emph{{Joe Biden (Q6279)}}'' is categorized as ``\emph{{Human (Q5)}}'' in Wikidata and linked to ``\emph{{Scranton (Q271395)}}'' with relation ``\emph{{birth place (P19)}}'', where prefix \emph{{Q}} and \emph{{P}} denote unique identities for entity and relation respectively in Wikidata\footnote{Wikidata links \url{https://www.wikidata.org/wiki/Q5},  \url{https://www.wikidata.org/wiki/Property:P19}}. As ontological KGs are well organized and canonicalized, one can efficiently query information and extract knowledge from them to assist NLP models in various tasks \cite{kb3,kb4,kb1,kb2,kepler,DualMatch,ClusterEA,Liu2022HiUREHE}. However, building and maintaining an accurate ontological KG requires large human effort \cite{Frber2015ACS}.

In contrast, open knowledge graphs such as ReVerb \cite{ReVerb2011} and OLLIE \cite{ollie-emnlp12} are built using {\it (subject, relation, object)} triples automatically extracted from millions of raw text by OpenIE systems \cite{stanford-openie,ReVerb2011,ollie-emnlp12}. They are frequently used to assist in building ontological KGs \cite{openie4kg1, openie4kg2} and slot filling \cite{openie4slot}. 
As OpenIE systems do not rely on pre-defined ontologies or human supervision, the extracted triples contain noun phrases (NPs) and relation phrases (RPs) that are not canonicalized. Take the open KG triples shown in Figure \ref{fig:canonic}(b) as an example. The NP \emph{``Joseph Biden''} and \emph{``Biden''} both refer to the US president Joe Biden, but the open KG regards them as two different nodes because of their different surface forms. 
On the other hand, \emph{``was born in''} in the first and second triple means \emph{``birth place of''} and \emph{``birth time of''} respectively, but the open KG cannot disambiguate them. These examples reveal the redundancy and ambiguity of uncanonicalized open KG \cite{vashishth2018cesi}, which makes querying open KG inaccurate and inefficient. To this end, open KG canonicalization aims to improve the quality of open KGs to 
the level of ontological KGs. It is therefore different from tasks such as entity linking \cite{kb3} and KB aligning \cite{elsahar2018t} that align entity mentions or sentences to an existing ontological KG.

Existing open KG canonicalization datasets such as ReVerb-base, ReVerb-ambiguous \cite{reverbbase}, ReVerb45K \cite{vashishth2018cesi} and CanonicNELL \cite{cuva} mainly focus on entity-level canonicalization of NPs, providing the gold \textbf{E}ntity-level \textbf{NP} \textbf{C}anonicalization (\textbf{NPC-E}). The blue tags and dashed boxes in Figure \ref{fig:canonic}(b) show examples of \textbf{NPC-E}, e.g., \emph{``Biden''} and \emph{``Joseph Biden''} should be canonicalized as the same entity \emph{Q6279}. However, these datasets do not provide the gold \textbf{RP} \textbf{C}anonicalization (\textbf{RPC}), and do not consider the \textbf{O}ntology-level \textbf{C}anonicalization of \textbf{NP} (\textbf{NPC-O}). \textbf{RPC} is to canonicalize RPs that mean same relation together, for example, the second and the third \emph{``was born in''} in Figure \ref{fig:canonic}(b) should be canonicalized into the same cluster of \emph{birth place (P19)}, different from the first one which means \emph{``birth time (P569)''}. Similarly, \textbf{NPC-O} is to canonicalize NPs that have same type together, for example, the \emph{``Scranton''} should be canonicalized into class \emph{``CountySeat''} and into class \emph{``Local Government''} together with \emph{``Atlantic County''}, it can be viewed as canonicalizing special ontological relations such as ``\emph{instance of}'', ``\emph{subclass of}'' represented by dotted arrows in Fig. \ref{fig:canonic}(a). We formally define these tasks in Sec. \ref{sec:define}.

\textbf{RPC} and \textbf{NPC-E} are important as parts of a canonicalization benchmark (1) Relations and ontology are 
necessary for an expressive KG \cite{klyne2004resource} (2) Most KG queries involve relations and ontology (e.g., the query \emph{``actress that was born in California''}, involve the relational constraint ``\emph{X, birth place, California}'' and the ontological constraint ``\emph{X, instance of, Actress}'').

In this paper, we present \datasetname, a complete benchmark for open KG canonicalization consisting of three subtasks: besides \textbf{NPC-E} which has been adequately studied in previous work, we additionally provide gold \textbf{RPC} and \textbf{NPC-O} along with their evaluation metrics. Gold \textbf{NPC-O} is obtained by querying the Wikidata using SPARQL, and \textbf{RPC} is obtained by performing Stanford OpenIE on sentences from Wiki20 (a distantly labeled relation extraction dataset), and a per-instance human revision process to ensure the quality of extracted RPs. We introduce the data construction process detailedly in Sec. \ref{sec:construction}. 

Our new benchmark makes it possible for the first time to quantitatively evaluate the full range of open KG canonicalization. We conduct comprehensive experiments to compare existing canonicalization methods as well as a few simple baseline methods proposed by us. Somewhat surprisingly, none of the existing methods utilizes pretrained contextualized word embedding, probably because previous work only focuses on NPC-E and NPs are often not very ambiguous, making contextualization not so helpful. For example, the ``\textit{Joe Biden}'' and ``\textit{Joseph Biden}''.
However, contexts are more helpful in RPC and NPC-O. For RPC, relations are more ambiguous and diverse in surface forms (e.g., ``\textit{was born in}'' in Figure \ref{fig:canonic}) and contexts are needed for disambiguation. For NPC-O, the RP and the other NP in the triple will help understand the type of an NP.
Therefore, our proposed baseline methods are based on pretrained language models (PLM) \cite{devlin2018bert,liu2019roberta,sun2019ernie} which produce contextualized embedding and have been shown to contain a certain amount of factual knowledge \cite{petroni-etal-2019-language,lauscher-etal-2020-common}. We found that, after properly encoding triples and contexts, our baseline methods outperform well on all three subtasks compared with previous state-of-the-art methods, especially on RPC and NPC-O. We also propose a triple-based pretraining method and find that it further boosts the performance on all subtasks. Therefore, our work provides strong baselines for future research on open KG canonicalization.

\begin{table*}[t]
\centering
\scalebox{0.82}{
\begin{tabular}{l||c|c|c|c|c|c|c|c} \toprule
                 & \# \textbf{NP} & \# \textbf{NPC-E} & \# \textbf{RP} & \# \textbf{RPC} & \# \textbf{NPC-O} & \# Triples & Avg triple len & Context (\%) \\ \hline
ReVerb-Base      & 290   & 150                        & 3K    & \xmark         & \xmark                        & 9K         &   5.26                    & 78\%              \\
ReVerb-Ambiguous & 717   & 446                        & 11K   & \xmark         & \xmark                        & 37K        &   5.27                   & 78\%              \\
ReVerb45K        & 15.5K & 7.5K                       & 22K   & \xmark         & \xmark                        & 45K        &  6.17                     & 91\%                   \\
CanonicNELL      & 8.7K  & 1.4K                       & 139   & \xmark         & \xmark                        & 20K        & 6.38                  & \xmark               \\ \hline \hline
\datasetname\ (Ours)           & 16.5K & 13.8K                      & 3.2K  & 79            & 2946                         & 18K        & 8.12                  & 100\% \\ \bottomrule           
\end{tabular}}
\caption{Statistics and comparison of open KG canonicalization datasets including ours.  \xmark \ means not available in the dataset (i.e., zero). ``Avg triple len'' is the average number of words in the triple. The last column shows the ratios of triples containing additional context in their source sentences. }
\label{tab:stat}
\end{table*}

In summary, our contributions are threefold. First, we propose a complete definition of the open KG canonicalization problem along with the metrics. Second, we construct the complete benchmark for open KG canonicalization consisting of entity-level and ontology-level NP canonicalization and RP canonicalization. Third, we propose a stronger baseline based on autoencoding PLMs and conduct a comprehensive empirical comparison of canonicalization methods on our benchmark.

\section{Open KG Canonicalization Datasets}
We introduce existing open KG canonicalization datasets and \datasetname. The statistics of datasets are shown in Table \ref{tab:stat}. 
\\
\textbf{ReVerb-Base} \cite{reverbbase}\ \  Constructed using the ReVerb open KB. As half of the NPs in ReVerb triples are linked to an entity in the ontological database FreeBase \cite{FreeBase}, the authors sample 150 FreeBase entities that have at least two surface forms, collect all triples containing these 150 entities, and use the entity labels as the gold NP clusters. \\
\textbf{ReVerb-Ambiguous} \cite{reverbbase} \ \  ReVerb-Ambiguous is constructed similarly as ReVerb-Base, it has 37K triples, but with only 445 gold NP clusters (entities). One problem with the ReVerb-Base and ReVerb-Ambiguous datasets is they contain too few NP clusters and too many NP aliases, which is inconsistent with real open KGs. \\ 
\textbf{ReVerb45K}\cite{vashishth2018cesi} \ \  ReVerb45K increases the entity number to 7.5K and has 45K triples in total. ReVerb45K, Reverb-Base, and ReVerb-Ambiguous extract a source sentence for each triple from ClueWeb09 \citet{callan2009clueweb09}. However, some of the source sentences are simply the concatenation of triples. \\
\textbf{CanonicNELL}\ \cite{cuva} \ \ Constructed using the open KB NELL \cite{mitchell2014never} and the entity linking information for NPs \cite{pujara2013knowledge}. They remove triples containing NPs without aliases. CanonicNELL does not provide source sentences. \\
\datasetname \ \textbf{(Ours)} \ \ As shown in the Table \ref{tab:stat}, the main differences of our dataset between others are that we additional provide gold RP canonicalization and ontology-level NP canonicalization. Constructed based on the large Ontological KG Wikidata\footnote{The official suggested replacement of Freebase after it retired \url{shorturl.at/kmnBR}}, the OpenIE system, a relation extraction dataset Wiki20m and human revisions, as detailed in next section. Our dataset contains 18K triples with their source sentences and we provide gold NPC-E, RPC and NPC-O annotations. We compare \datasetname \ with existing datasets in Table \ref{tab:stat}. Although our dataset is middle-sized, it has the longest average triple length and the largest number of unique NPs, indicating the diversity of the surface forms of NPs and RPs. Providing source sentences of extracted OpenIE triples is natural but important since additional contextual information can be helpful in understanding and disambiguating NPs and RPs. We ensure all triples contain rich context, and the average length of source sentences is 21. We show some data samples in Appendix \ref{app:example}, and analyze our data in Sec. \ref{sec:construction}.
\section{Task Definition and Evaluation Metrics}
\label{sec:define}
\paragraph{Task Definition\ } The goal of open KG canonicalization is to assign NPs and RPs in triples into clusters, such that NPs that refer to the same entity (NPC-E) or have the same type (NPC-O) are clustered together, and similarly, RPs that refer to the same relation are clustered together. Note that the task is unsupervised, meaning that the canonicalizer does not have access to gold annotations. 
We have $N$ samples containing triples and their corresponding source sentences: 
$\mathcal{T} = \{ c_i, t_i=(s_i, r_i, o_i)  \vert \  i=1 \dotsc N\} $, where
$c_i$ is the $i$-th sentence, $t_i$ is the $i$-th triple containing subject NP $s_i$, RP $r_i$, and object NP $o_i$. 
$\mathcal{S}=\{(s_i,i) \vert \ i=1 \dotsc N\}$ is the indexed subject NP set. The indexed RP set $\mathcal{R}$ and object NP set $\mathcal{O}$ are defined similarly as $\mathcal{R}=\{(r_i,i) \vert \ i=1 \dotsc N\}$ and $\mathcal{O}=\{(o_i,i) \vert \ i=1 \dotsc N\}$. We have $\vert \mathcal{S} \vert = \vert \mathcal{O} \vert = \vert \mathcal{R} \vert = N$.
The gold NPC-E, RPC, and NPC-O annotations are defined as sets of clusters. As subject NPs and object NPs are asymmetric \cite{juffs1995parsing, mcginnis2002object}, we follow \citet{vashishth2018cesi} and evaluate the clusters of subject NPs and object NPs separately. The gold NPC-E and NPC-O for subject NPs are defined as $\textit{NPC-E (Subj)} = \{C_1 \dotsc C_{K^E_s} \}$, $\textit{NPC-O (Subj)} = \{C_1 \dotsc C_{K^O_s} \}$, where $C_i$ denotes the $i$-th cluster of NP. The NPC-E and NPC-O of object NPs are defined similarly, the gold RPC is defined as $\textit{RPC}$ as $\{C_1 \dotsc C_{K_r}\}$.
\textit{NPC-E (Subj)} is a \emph{non-overlapping} cluster assignment and satisfies two conditions: (1) $\bigcup_{i=1}^{K_s^E} C^i =\mathcal{S}$; (2) $C_i \cap C_j=\emptyset, i\not=j$. \textit{NPC-E (Obj)} and \textit{RPC}  satisfy similar conditions. \textit{NPC-O (Subj)} and \textit{NPC-O (Obj)} are \emph{overlapping} cluster assignments, i.e., we allow an NP to belong to multiple clusters, so they only need to satisfy the first condition. The task is to predict the cluster assignments of NPs and RPs given their source triples and sentences. Following previous works, we assume the cluster number is unknown beforehand and split our data into the dev (20\%) and test (80\%) sets. \\
\paragraph{Task Evaluation\ } Most clustering algorithms such as K-means \cite{lloyd1982kmeans} and Hierarchical Agglomerative Clustering (HAC) \cite{maimon2005hac} produce non-overlapping cluster assignments, and several algorithms (e.g., HAC) can also produce hierarchical and overlapping cluster assignment. 
For the \textit{NPC-E} subtask, we adopt the classic macro, micro and pairwise metrics to compare the gold and predicted \textit{NPC-E} cluster assignments (please refer to App. \ref{app:metric} for details).
For \textit{RPC}, the macro metrics that calculate the fractions of pure clusters are too strict because gold RP clusters are large and hence are unlikely to be pure. Therefore we only use the micro and pairwise metrics to evaluate \textit{RPC}.
\begin{table}[t]
\scalebox{0.7}{
\begin{tabular}{c|ccc} \toprule
                             & \emph{NPC-E}                  & \emph{RPC}             & \emph{NPC-O}                            \\ \midrule
Gold     & non-overlapping         & non-overlapping  & overlapping                       \\ \hline 
Predicted & non-overlapping         & non-overlapping  & \makecell[c]{ non-overlapping / \\ overlapping}      \\ \hline \hline
\textbf{Metric}   & Ma, Mi, Pair & Mi, Pair & \makecell[c]{ Ma, Mi, Pair / \\ $J_{g\to p}, J_{p\to g}$ } \\ \bottomrule
\end{tabular}}
\caption{Evaluation of the three subtasks. Ma, Mi, Pair are abbreviations of macro, micro and pairwise metrics. }
\label{tab:metric}
\end{table}

For \textit{NPC-O}, the gold cluster assignments are overlapping. If the predicted clusters are non-overlapping, we can apply the macro and pairwise metrics and a modified micro metric (Appendix \ref{app:mi}). If the predicted clusters are overlapping, say $\mathcal{P}=\{C^p_1 \dotsc C^p_M\}$,
we propose evaluation metrics $J_{g\to p}$ and $J_{p\to g}$ based on the Jaccard index \cite{jaccard1908nouvelles,tanimoto1958elementary}. $J_{g\to p}$ (Eq.\ref{eqa:1}) calculates the average Jaccard index of a gold cluster and its best matched predicted cluster. $J_{p\to g}$ is similarly defined but with the roles of \textit{NPC-O} and $\mathcal{P}$ switched. Table \ref{tab:metric} summarizes the evaluation metrics of each subtask. 
\begin{equation}
\small{
\begin{aligned}
    & \text{Jaccard}(g,p) = \frac{\vert g \cap p \vert}{\vert g \cup p \vert}\\
    & J_{g\to p} =  \frac{1}{\vert \textit{NPC-O} \vert} \sum_{g \in \textit{NPC-O}} \max_{p \in \mathcal{P}} \big( \text{Jaccard}(g,p) \big)
\end{aligned}}
\label{eqa:1}
\end{equation}

\section{Construction of Our Dataset}
\label{sec:construction}
\begin{figure}[t]
    \centering
    \scalebox{0.20}{
    \includegraphics{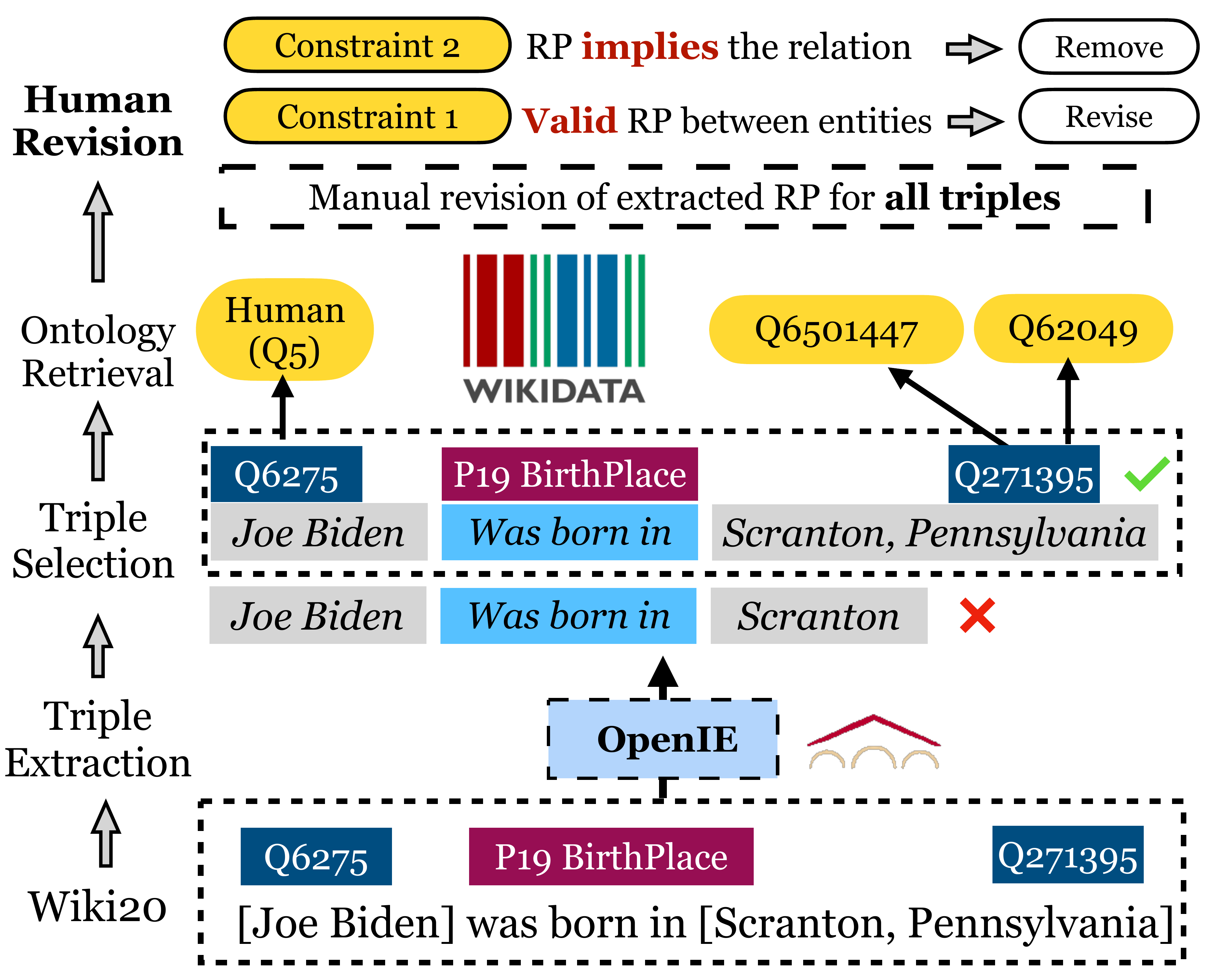}}
    \caption{Steps of dataset construction.}
    \label{fig:construction}
\end{figure}

We illustrate the construction process of \datasetname \ in Figure\ \ref{fig:construction}. 
We rely on the Wiki20 dataset \cite{han-etal-2020-data} to obtain the source sentence and the gold NPC-E. Wiki20 is a large multi-domain relation extraction dataset constructed by aligning the Wikipedia corpus with Wikidata using distant supervision. As shown in the bottom of Figure \ref{fig:construction}, each sample of Wiki20 contains a sentence with the object and subject NP spans labeled and linked to entities in Wikidata and the relation between them is also labeled. To ensure data quality, we use the recently revised version of Wiki20 \cite{gao-etal-2021-manual}, which aligns the Wiki20 relation labels with the supervisedly constructed Wiki80 dataset \cite{han2019opennre} and provides 56K human-annotated data samples. The object and subject NP spans and its entity linking information (e.g., \emph{Q6275}) are from Wikipedia and have high precision, so we directly use it for task NPC-E.

\paragraph{Extracting Relational Phrases} Wiki20 only provides the relation label of two NPs for each instance. We further extract RP for Wiki20 instances to obtain full open KG triples. We first discard samples with the relation label ``NA'' and then run the Stanford OpenIE system on Wiki20 sentences to extract triples. We choose Stanford OpenIE\footnote{\url{https://stanfordnlp.github.io/CoreNLP/}} because compared with older OpenIE systems such as ReVerb and NELL that are used in constructing previous datasets, Stanford OpenIE can leverage the linguistic structure of a sentence and generalizes better to out-of-domain and longer utterances \cite{stanford-openie}. We empirically find that Stanford OpenIE yields a better recall and can extract more triples per sentence compared to ReVerb. We use the default model configuration of Stanford OpenIE. After obtaining the OpenIE triples of each non-NA Wiki20 instance, we select the triples whose subject NP and object NP are consistent with the NP spans provided by Wiki20. This triple selection step ensures the NPs in the extracted triples have gold NPC-E annotations, and remove wrong relation spans caused by wrongly extracted head and tail entities. We filter out $88\%$ of the original triples through this step. Although this step reduces noises caused by OpenIE, the extracted relation spans could still be wrong in two ways: 
\begin{enumerate}[leftmargin=0.5cm]
    \item \textbf{Invalid RP between correct NPs}. For example, for sentence \emph{``\dots the Althing, the ruling legislative body of Iceland \dots''}, OpenIE wrongly extracts \emph{(the Althing, body of, Iceland)}, while the true triple should be \emph{(the Althing, {\color{magenta} ruling legislative body of}, Iceland)}.
    \item \textbf{Correct NPs and valid RP but RP does not imply the relation given by Wiki20}. For the given relation \emph{mother of} and sentence \emph{``\dots bart and lisa got sent out of the house by marge simpson \dots''}, the extracted triple \emph{(lisa, {\color{magenta} got sent out of the house by}, marge simpson)} is valid but cannot imply the \emph{mother of} relation.
\end{enumerate}
Therefore, 
we manually check all the extracted triples for these two types of errors, correcting invalid relational phrase spans and removing triples whose RP cannot imply the given relation. We also standardize the form of RP (e.g., OpenIE sometimes includes ``a'' and ``the'' and sometimes does not). The detailed guidelines for the check and revision process are shown in the Appendix \ref{app:guide_rp}. The error analysis is shown in Table \ref{tab:error}.
\begin{table}[h]
\centering
\scalebox{0.8}{
\begin{tabular}{ll} \\ \toprule
  \textbf{Error Type}          & \textbf{Rate} \\ \midrule
Invalid RP         &   23.5\%           \\
RP doesn't imply relation &   5.5\%          \\ \bottomrule  
\end{tabular}}
\caption{OpenIE error analysis.}
\label{tab:error}
\end{table}

\begin{figure}[h]
    \centering
    \scalebox{0.3}{
    \includegraphics{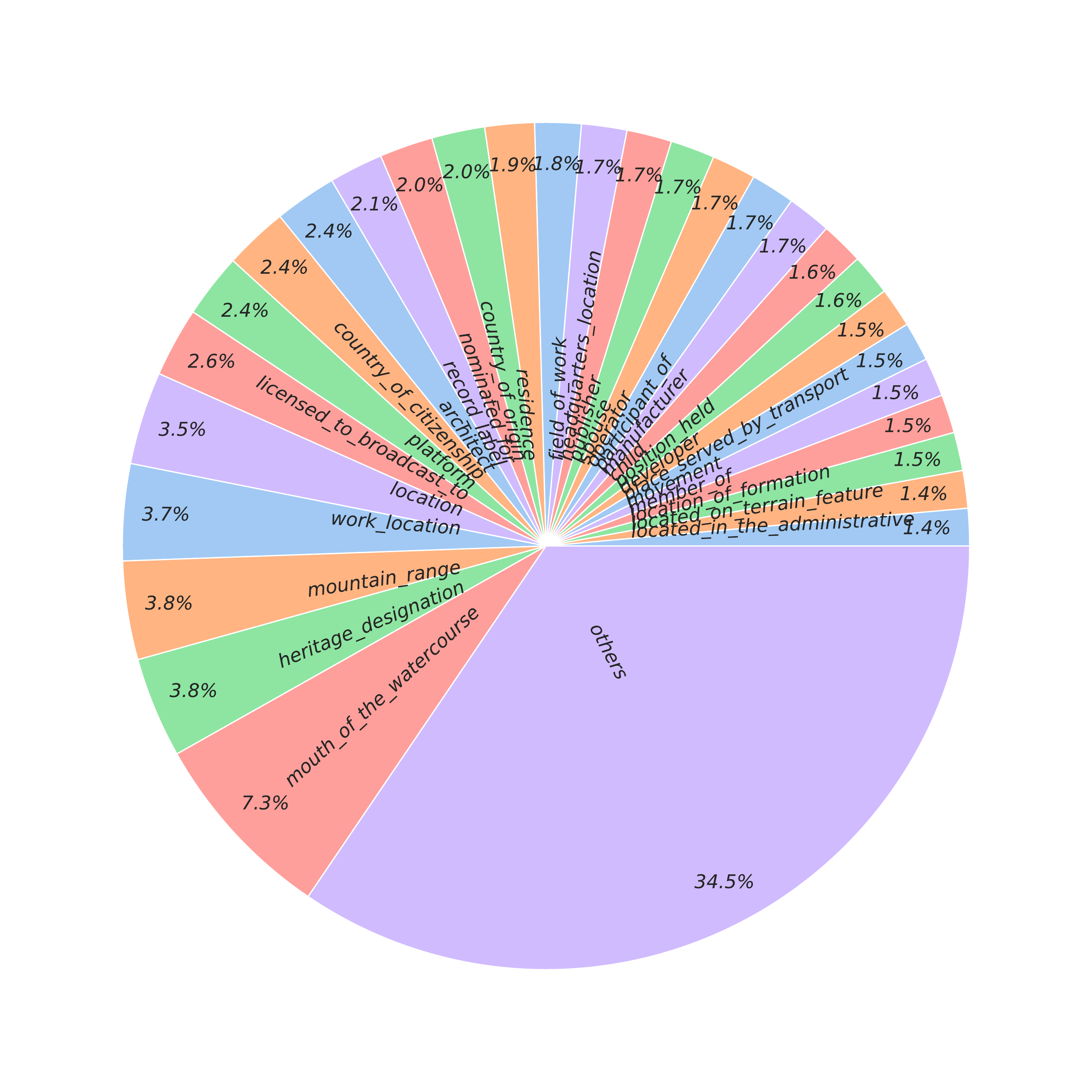}}
    \caption{Pie charts of 79 RP clusters.}
    \label{fig:RPC}
\end{figure}

After all these steps, we obtain an open KG consisting of 18K triples. Similar to NPC-E, we use the relation labels given by the Wiki20 annotations (e.g., \emph{P19}) as the gold RPC. As shown in Figure \ref{fig:RPC}, the constructed open KG contains 79 relations in various domains, such as relations between geopolitical entities ({\it mouth of the watercourse} ($7.3\%$), {\it mountain range} ($3.8\%$), etc), relations between people ({\it spouse of} ($1.7\%$), {\it child of} ($1.6\%$), etc), and various relations between people and other objects ({\it citizenship} ($2.4\%$), {\it work location} ($3.7\%$), etc). The extracted RPs are diverse in surface forms. The number of distinct RPs is 3.2K. We show RP examples in Table \ref{tab:rp_example1}. There exist some RPs that represent multiple relations and one representative example is \emph{``in''}.

\begin{table}[]
\centering
\scalebox{0.85}{
\begin{tabular}{c} \\ \toprule
\textbf{\emph{spouse of}}                                   \\ 
\emph{\thead{was married twice to , was married to, \\
lover, consort of, second husband, widow of\\
's wife, 's second wife, arranged a wedding with } }\\ \midrule
\textbf{\emph{mountain range}}                        \\ 
\emph{\thead{peak, large nunatak, summits of, the only crossing of the  \\
most prominent feature of, small glacier, summits in\\
valley in, only crossing of, northernmost subrange of, \textbf{in}}}   \\ \midrule
\textbf{\emph{location}}              \\ 
\emph{\thead{is headquartered in, moved to,  is carved on  \\
took place at, ironworks in, was again held at, \textbf{in}}} \\ \bottomrule 
\end{tabular}}
\caption{RP examples.}
\label{tab:rp_example1}
\end{table}

\begin{figure}[h]
    \centering
    \scalebox{0.38}{
    \includegraphics{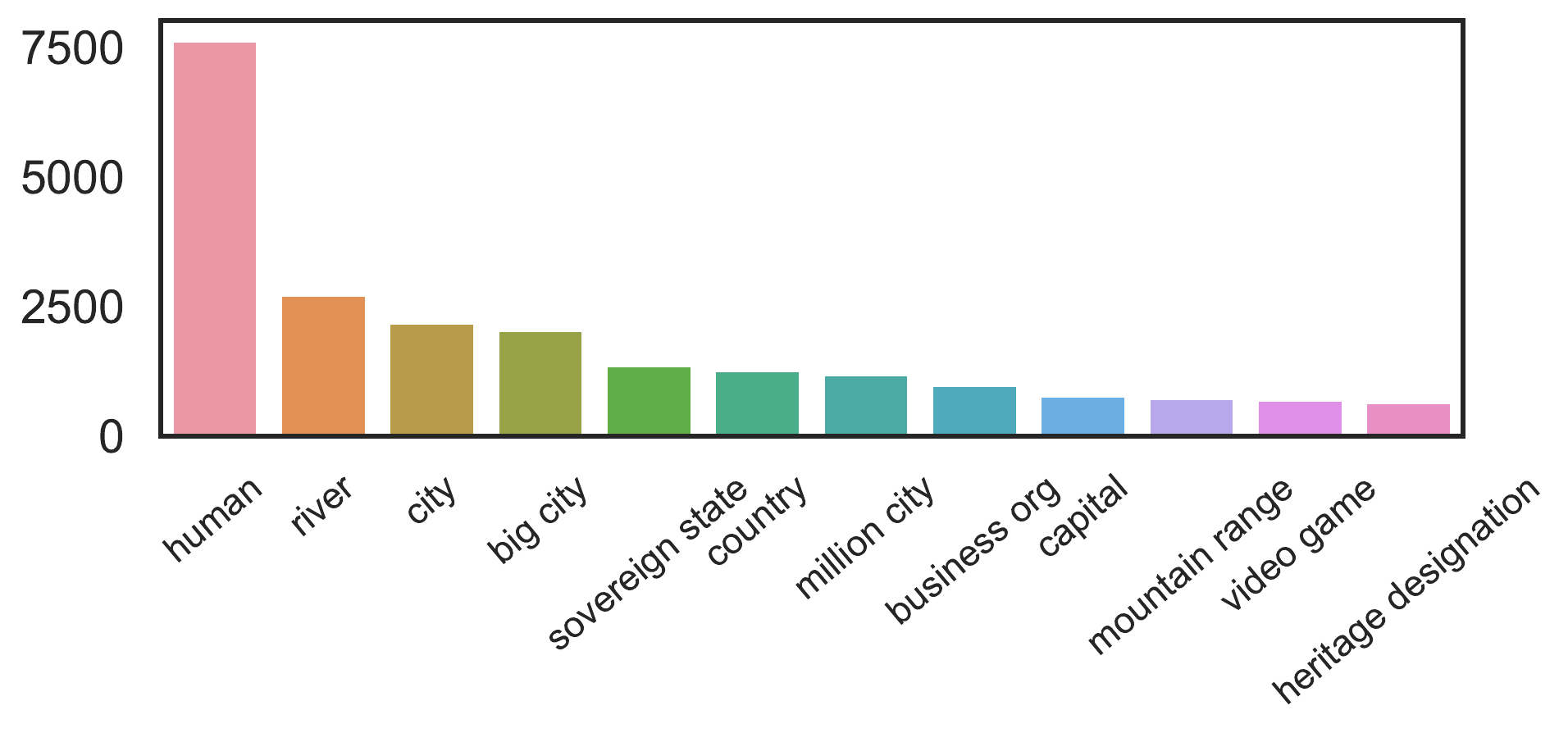}}
    \caption{Size of top 12 ontological NP clusters.}
    \label{fig:NPC-O}
\end{figure}

\begin{figure}[h]
    \centering
    \scalebox{0.23}{
    \includegraphics{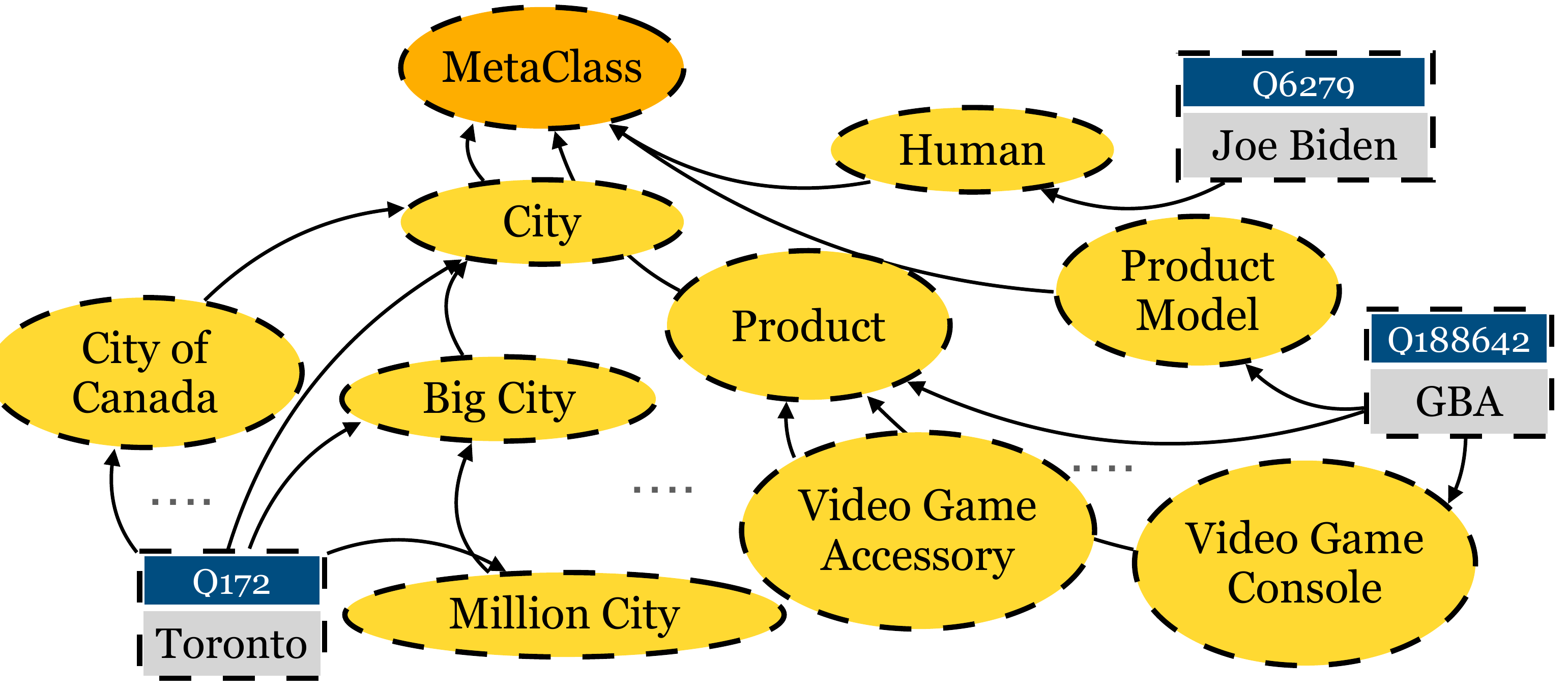}}
    \caption{Part of the class hierarchy in our dataset.}
    \label{fig:hierarchy}
\end{figure}

\paragraph{Extracting Ontology}
To obtain ontology-level NP clusters for the NPC-O subtask, we query Wikidata for the classes of each entity. For example, to obtain the classes of \textit{``Joe Biden (Q6275)''}, we run the SPARQL (RDF query language) query `` {\small {\tt Q6275   P31   ?}}'', where 
P31 represents the ``instance of'' relation 
in Wikidata. This query obtains all the classes of an NP. If an NP does not have a class, its NPC-O annotation is the same as its NPC-E annotation. If an NP has more than one class, we include all of them in the NPC-O annotation (e.g., {\it city} and {\it big city} for ``New York''). We query Wikidata using a third-party client Wikidata Integrator\footnote{\url{https://github.com/SuLab/WikidataIntegrator}}.
As the ontology information in Wikidata is crowdsourced and contains errors, we apply pattern-based corrections to the extracted ontological NP clusters, for example, if an NP belongs to the cluster {\it million cities}, it should also belong to the cluster {\it city}. The resulting 2.9K ontological NP clusters form a 6-level overlapping hierarchy which allows a node to have more than one parent. We illustrate part of the hierarchy in Figure \ref{fig:hierarchy} and show the statistics of the top 12 ontological NP clusters in Figure \ref{fig:NPC-O}. 

\section{Comprehensive Evaluation of Methods}
Our benchmark makes it possible to conduct a comprehensive empirical comparison of different methods on the full range of open KG canonicalization. Below we first give an overview of existing methods and propose a few new baseline methods. Then we present our experimental settings and results.

\subsection{Previous Methods}
\textbf{Non-neural Methods\ } \citet{reverbbase} utilizes token features such as TF-IDF scores and Jaccard token similarity to canonicalize NPs. They merge similar NPs based on a threshold (tuned on the validation set) to form clusters. For RPs, they use AMIE \cite{galarraga2013amie}, an unsupervised algorithm based on statistical rule mining, to obtain relation clusters. \citet{vashishth2018cesi} use additional side information obtained from various sources (such as PPDB \cite{PPDB}, WordNet \cite{WordNet}) to produce clusters. \\
\textbf{SE-HAC\ } Trivial baseline that performs HAC clustering over phrase embeddings produced by averaging static word embeddings such as GloVe \cite{pennington2014glove} or random embeddings. \\
\textbf{CESI\ } \citet{vashishth2018cesi} encode phrases using the same method as in SE-HAC; then apply the HolE graph embedding algorithm \cite{nickel2016holographic} on triples and incorporate side information to finetune the embedding, and finally run HAC clustering on the learned embeddings. \\ 
\textbf{CUVA\ } \citet{cuva} adopt VAEGMM \cite{vaegmm} to jointly learn and cluster the embeddings. They initialize VAEGMM by performing HAC clustering on GloVe NP and RP embeddings, and then simultaneously optimize the knowledge embedding loss, side information loss, and VAE loss for the final clustering.

Previous methods encode NPs and RPs using either token frequency features or static word embedding. Although CESI and CUVA learn graph embedding on open KG triples, they assign the same representation to phrases with the same surface form and therefore cannot deal with ambiguity. No method utilizes original sentences to provide additional contexts. HAC is a popular choice of the clustering algorithm because it does not require knowing the number of clusters, but instead requires a distance threshold indicating when to stop merging. Unlike the number of clusters, the threshold can be tuned on a validation set and directly applied to the test set.

\subsection{PLM-Based Baseline Methods}
We propose a set of new baseline methods based on PLMs that produce contextualized embedding.
\begin{figure}
    \centering
    \scalebox{0.25}{
    \includegraphics{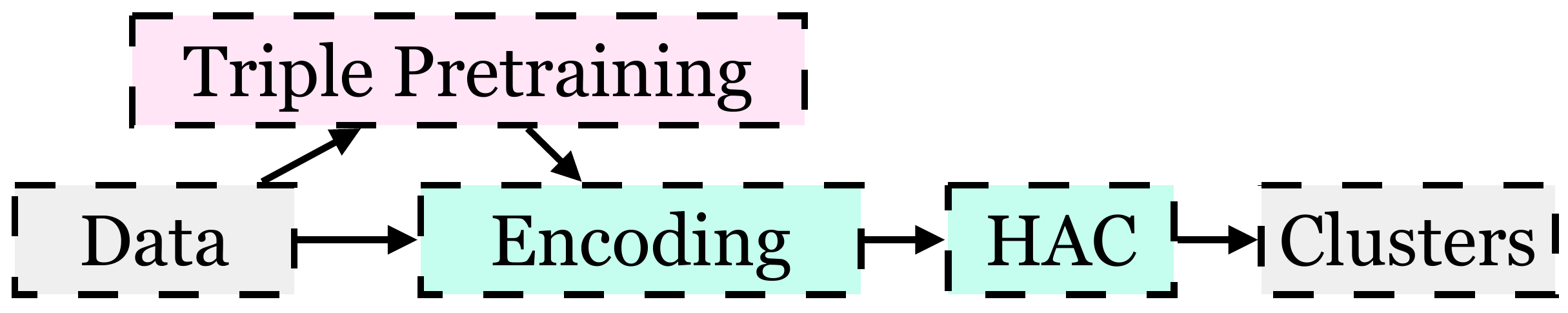}}
    \caption{Pipeline of the proposed PLM-based method.}
    \label{fig:pipline}
\end{figure}
We use a pipeline similar to CESI as shown in Figure \ref{fig:pipline}. We encode NPs and RPs using different PLMs, PLM layers, and span representation methods and apply HAC clustering over their representations. We use the cosine similarity as the distance function and apply the complete linkage variant of HAC clustering because we prefer compact clusters and the single linkage variant suffers from the chaining phenomenon. Before encoding, an optional triple-level continuous pretraining step can be applied for better canonicalization. Token similarity and other side information are not used in our PLM-based method, but we generate them for our data using the code provided by \citet{vashishth2018cesi} to facilitate running of other methods.

\subsubsection{Encoding}
\paragraph{PLMs} We use autoencoding PLMs\footnote{Huggingface models\ \url{https://huggingface.co/}} including BERT \cite{devlin2018bert}, RoBERTa \cite{liu2019roberta}, ERNIE2.0 \cite{sun2019ernie} which integrates entity information, and SpanBERT \cite{joshi-etal-2020-spanbert} which is pretrained with span masks.
\paragraph{Input} Given a triple $t_i=(s_i, r_i, o_i)$ and its corresponding source sentence $c_i$, we formulate the input of PLM in the following four ways to obtain the contextualized embedding of words in the NPs and RP. Note that the fourth method \textit{sep} independently encodes each phrase in the triple.
\begin{equation}
\small{
    \begin{aligned}
        \textit{sentence: \ } & [CLS]\ \dots s_i \dots r_i \dots o_i \dots \ [SEP] \\
        \textit{triple: \ } & [CLS]\ s_i\ r_i\ o_i\ [SEP] \\
        \textit{triple-sep: \ } & [CLS]\ s_i\  [SEP]\ r_i\ [SEP]\ o_i\ [SEP] \\
        \textit{sep: \ } & [CLS]\ s_i/r_i/o_i\ [SEP] \\
    \end{aligned}}
    \label{eq:2}
\end{equation}
\paragraph{Phrase Representation} After obtaining the contextualized embedding of words in an NP or RP span, denoted as $h_i\dotsc h_j$, we follow \citet{toshniwal2020cross} and use three methods to produce a single span embedding. Following \citet{timkey2021all}, we also standardize the embeddings to remove rogue dimensions (Appendix \ref{app:std}). Previous work \cite{vulic2020probing, liu-etal-2019-linguistic} shows that different layers of a PLM contain different information, so we investigate contextualized embedding from different layers .
\begin{equation}
\small{
    \begin{aligned}
        \textit{mean: \ } & average\_pooling(h_i\dotsc h_j)\\
        \textit{max: \ } & max_-pooling(h_i\dotsc h_j) \\
        \textit{diff-sum: \ } & [h_i-h_j; h_i+h_j] \\
    \end{aligned}}
    \label{eq:3}
\end{equation}

\subsubsection{Triple-level Pretraining}
Inspired by the HolE algorithm used in previous works \cite{vashishth2018cesi, cuva}, we may perform an optional triple-level continuous pretraining step before encoding to mimic the link prediction objectives in knowledge graph embedding learning. For each sentence in our dataset, we randomly mask a phrase in the triple and then train the PLM to predict the whole masked span. We perform pretraining for 10 epochs using the AdamW optimizer \cite{loshchilov2018decoupled} with a linear scheduler and a start learning rate of 5e-5. We then use the continuously pretrained version of PLM as the phrase encoder. We also use the causal subword-level MLM strategy in BERT \cite{devlin2018bert} for comparison.

\subsection{Experimental Setup}
\begin{table*}[h]
\centering
\scalebox{0.68}{
\begin{tabular}{l|cc|c|cccc} \toprule
                              & \multicolumn{2}{c|}{\textit{\textbf{NPC-E}}}    & \textit{\textbf{RPC}}      & \multicolumn{4}{c}{\textit{\textbf{NPC-O}}}                                                                     \\
\multirow{-2}{*}{}            & \textit{\textbf{Subj}} & \textit{\textbf{Obj}} & \textit{\textbf{Relation}} & \textit{\textbf{Subj}} & \textit{\textbf{Subj-Jaccard}} & \textit{\textbf{Obj}} & \textit{\textbf{Obj-Jaccard}} \\ \midrule

\textit{Token Sim+SI} \cite{reverbbase}                & 82.90                  & 79.35                 & 33.94                      & 33.14                  &  -                             & 40.59                 & -                              \\
\textit{Random+HAC}                                    & \textbf{85.32}                  & 85.11                 & 35.98                      & 37.31                  & 38.79                          & 44.90                 & 44.40                         \\
\textit{GloVe+HAC}                                     & 78.31                  & 86.57                 & 35.57                      & 38.00                  & 39.04                          & 47.35                 & 45.45                         \\
\textit{GloVe+HolE+HAC} \cite{vashishth2018cesi}       & 80.13                  & 87.33                 & 17.93                      & 39.63                  & 39.87                          & 48.25                 & 47.18                         \\
\textit{GloVe+SI+HAC} \cite{vashishth2018cesi}         & 80.42                  & \textbf{87.34}                 & 17.91                      & 39.87                  & 39.83                          & \textbf{48.26}                 & 47.11                              \\
\textit{CESI} \cite{vashishth2018cesi}                 & 80.11                  & 86.82                 & 18.37                      & \textbf{39.93}                  & \textbf{39.89}                                & 48.25                 & \textbf{47.19}                               \\
\textit{VAEGMM+SI} \cite{cuva}                         & 80.86                  & 82.91                 & 34.10                      & 37.41                  &  -                             & 46.90                 & -                              \\
\textit{VAEGMM+HolE} \cite{cuva}                       & 80.15                  & 82.87                 & 36.12                      & 37.22                  &  -                             & 46.88                 & -                              \\
\textit{CUVA} \cite{cuva}                              & 80.68                  & 82.95                 & \textbf{36.13}                      & 37.09                  &  -                             & 46.89                 & -                             \\ \hline \hline
\rowcolor[HTML]{EEF5FF} 
\textit{Bert-base}            & \textbf{86.93}                  & 86.91                 & 54.47                      & \textbf{42.97}                  & \textbf{44.16}                          & 50.71                 & 46.99                         \\
\rowcolor[HTML]{EEF5FF} 
\textit{Roberta-base}         & 82.85                  & 85.00                 & 41.08                      & 39.21                  & 41.24                          & 46.07                 & 44.14                         \\
\rowcolor[HTML]{EEF5FF} 
\textit{SpanBert-base}        & 84.38                  & 86.32                 & 44.04                      & 41.61                  & 43.16                          & 47.29                 & 45.35                         \\
\rowcolor[HTML]{EEF5FF} 
\textit{ERNIE2.0-base}        & 86.68                  & \textbf{88.11}                 & \textbf{54.66}                      & 42.60                  & 44.05                          & \textbf{52.27}        & \textbf{47.05}                         \\ \hline
\rowcolor[HTML]{EEF5FF} 
\textit{Bert-base-triple}     & 86.01                  & \textbf{88.92}                 & \textbf{58.45}             & \textbf{43.71}         & \textbf{45.19}                 & \textbf{51.78}                 & \textbf{47.49}                \\
\rowcolor[HTML]{EEF5FF} 
\textit{Roberta-base-triple}  & 85.37                  & 87.22                 & 50.81                      & 42.29                  & 44.33                          & 50.31                 & 46.81                         \\
\rowcolor[HTML]{EEF5FF} 
\textit{SpanBert-base-triple} & 85.73                  & 85.89                 & 46.18                      & 42.53                  & 44.23                          & 47.60                 & 45.56                         \\
\rowcolor[HTML]{EEF5FF} 
\textit{ERNIE2.0-base-triple} & \textbf{87.21}         & 86.93                 & 57.28                      & 43.21                  & 44.33                          & 50.66                 & 47.27                         \\ \hline \hline
\rowcolor[HTML]{EEF5FF} 
\textit{Bert-large}           & \textbf{87.09}                  & \textbf{89.05}        & \textbf{50.31}                      & 42.34                  & \textbf{44.16}                          & 50.71                 & \textbf{47.25}                         \\
\rowcolor[HTML]{EEF5FF} 
\textit{Roberta-large}        & 83.50                  & 85.81                 & 40.51                      & 39.88                  & 42.35                          & 48.35                 & 45.58                         \\
\rowcolor[HTML]{EEF5FF} 
\textit{SpanBert-large}       & 86.32                  & 86.67                 & 45.84                      & 40.96                  & 42.90                          & 47.92                 & 45.68                         \\
\rowcolor[HTML]{EEF5FF} 
\textit{ERNIE2.0-large}       & 86.21                  & 88.86                 & 49.80                      & \textbf{42.71}                  & 44.01                          & \textbf{51.98}                 & 47.22                         \\ \bottomrule 
\end{tabular}}
\caption{Averaged metrics (of Table \ref{tab:metric}) on all the subtasks. For example, \textit{\textbf{NPC-O, Subj}} is the average of Ma,Mi and Pair metrics on the ontology-level canonicalization of subject NPs, and \textit{\textbf{NPC-O, Obj-Jaccard}} is the average of $J_{p\to g}$ and $J_{g\to p}$ for the overlapping clustering assignment of object NPs. Full results including the results of \textit{large-triple} models are shown in Appendix \ref{app:fullres}}
\label{tab:res}
\end{table*}

For each subtask, we use grid search to tune the HAC distance threshold on the dev set to obtain non-overlapping clusters for all the methods. We select the best threshold based on the average of the metrics shown in Table \ref{tab:metric}. We obtain overlapping clusters for NPC-O from the full HAC hierarchy. As HAC is deterministic, we run experiments once for methods without randomness and four times for methods involving random initialization (CUVA, Random+HAC). 
As Token Sim+SI and VAEGMM based methods cannot provide overlapping cluster assignments, we do not evaluate them by metrics based on the Jaccard index.
For our PLM-based methods, we select the best input form and span representation strategy based on the dev set performance. We also compare different \textbf{encoding strategies} in Appendix \ref{app:encode}, and \textbf{layer-wise performances} in Appendix \ref{app:layers}. 
\subsection{Overall Results}
We report averaged metrics for each subtask in Table \ref{tab:res} because of limited space. The full results are shown in Appendix \ref{app:fullres}.
The results show that our PLM-based baseline methods outperform previous methods in most cases, especially on RPC and NPC-O, indicating the importance of contextual information. Trivial baselines such as Token Sim+SI, Random+HAC and GloVe+HAC already perform well (around 80\%) on NPC-E, because NPs referring to the same entity usually have similar surface forms and do not have to rely on contexts for correct prediction. However, they perform badly on RPC and NPC-O, because surface forms alone are no longer adequate for these two subtasks because of higher ambiguity.
CESI has bad RPC performance but is very competitive on NPC-E (Obj), and better than SpanBERT and RoBERTa without triple-level pretraining, but is still worse than the other PLM-based methods. CUVA performs generally badly, probably because it is sensitive to VAEGMM initialization and relies heavily on side information. As our dataset has the longest average triple length and consists of texts from various domains, it could be more challenging for methods that do not use contextualized embedding.

For PLM-based methods, BERT leads to the best overall performance on NPC-E (Obj), RPC and NPC-O (Subj); ERNIE2.0 performs best on NPC-E (Subj) and NPC-O (Obj) and is comparable to Bert on NPC-E (Obj) and RPC; RoBERTa and SpanBERT fall behind, but are still better than most other non-PLM methods on NPC-O and RPC. Large PLMs are better than base PLMs on NPC-E, comparable on NPC-O, but worse on RPC. We also found the triple-level pretraining effective, having a positive influence in most cases, especially on RPC (e.g., +9.73 for RoBERTa). In contrast, using the causal subword-level pretraining for Bert improves the object NPC but harms the subject NPC and RPC (-1.07 points). A detailed comparison between triple-level and subword-level pretraining is shown in Appendix \ref{app:triple}.
\section{Conclusion}
We present \datasetname, a complete benchmark for open KG canonicalization.  \datasetname\ consists of three subtasks, entity-level and ontology-level NP canonicalization, and RP canonicalization. We construct the data and propose the evaluation metrics for the RPC and NPC-O that are not been adequately studied before. We also propose a stronger canonicalization method based on autoencoding PLMs and conduct a comprehensive comparison of different canonicalization methods on our dataset.
 
For future study, NPC-O and RPC still have a lot of room for improvement and the efficiency of canonicalization methods is also worth studying. We also note that \datasetname\  can be additionally used as a probing benchmark for PLMs and as a phrase-level relation classification dataset.

\section{Acknowledgement}
This work was supported by the National Natural Science Foundation of China (61976139) and by Alibaba Group through Alibaba Innovative Research Program.

\section*{Limitations}
One limitation of our work is that, the size of our dataset (18K) is relatively small compared to previous datasets (Table \ref{tab:stat}). Another limitation is that, similar to previous work, we perform clustering for three subtasks and evaluate the canonicalization results independently, but canonicalization of the head NP, tail NP and RP is in fact closely correlated. For example, the NPC-O clusters of the head NP and tail NP reveal the domain and range of the relation given by RPC. We leave jointly canonicalization and evaluation as future work. Our proposed baseline is straightforward. We encourage future studies to investigate better canonicalization methods based on pretrained language models.

\section*{Ethics Statement}
Our dataset is constructed based on Wiki20 and Wikidata. The two sources are both publicly available. Wiki20 is under the MIT Licence and the Wikidata is under the Creative Commons CC0 License. Both of them allow modification and distribution. Regarding human revision during dataset construction, the annotators were properly paid. The annotating procedure lasted 12 days and the daily workload was relatively light: around 2.5 hours per day. During human inspection, we did not identify any unethical instances in our dataset. Regarding baseline models, we use PLMs as our text encoder and our task is inherently unsupervised. As PLMs are learned on large corpora, our method can potentially create biased clustering results. How to de-bias PLM embedding is worth further investigation.

\bibliography{anthology,main}
\bibliographystyle{acl_natbib}

\appendix

\section{Dataset Examples}
\label{app:example}

\begin{figure*}
    \centering
    \scalebox{0.3}{
    \includegraphics{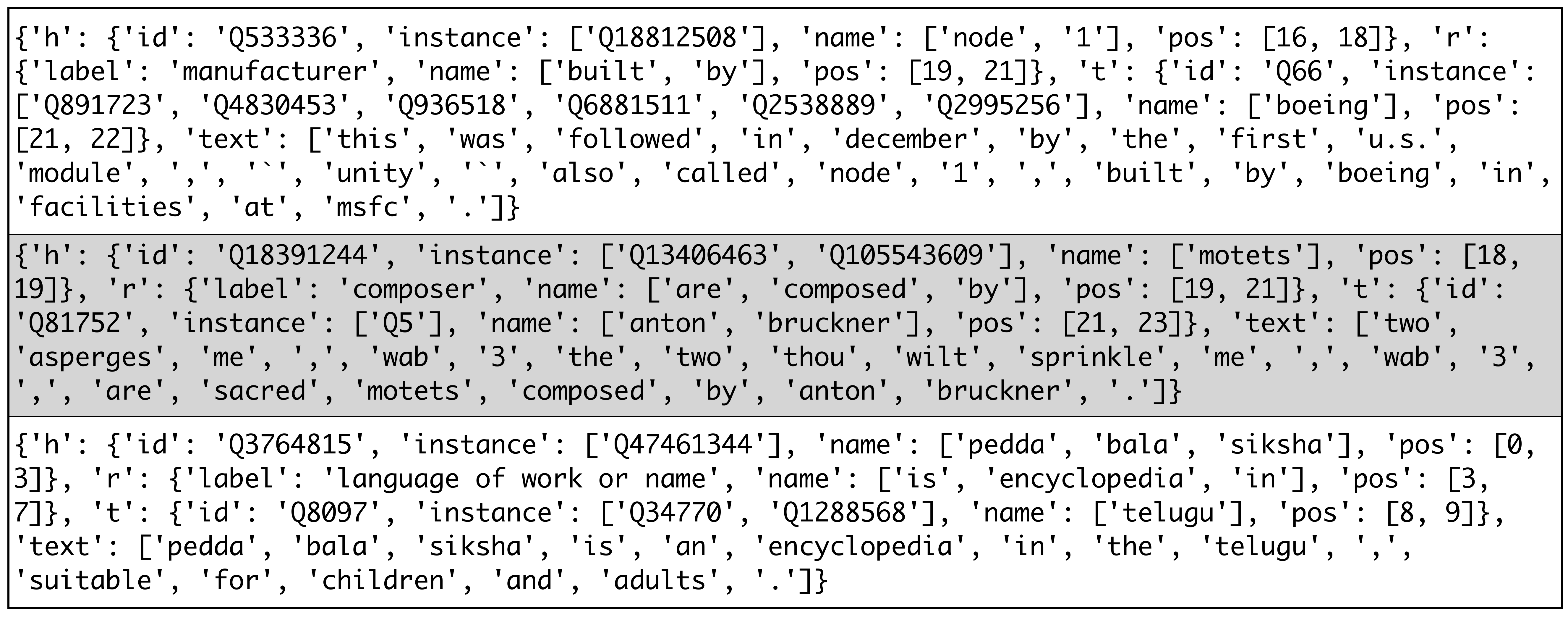}}
    \caption{Examples of our dataset, ``h'' means head or subject NP, ``r'' means relation, ``t'' means tail or object NP. ``instance'' stands for the gold ontology-level clusters. }
    \label{fig:example}
\end{figure*}

We show examples of our dataset in Figure \ref{fig:example}.

\section{Guidelines for Revising Relational Phrases}
\label{app:guide_rp}

\subsection{RP Annotating Procedure}
\label{app:overall1}
\begin{enumerate}[leftmargin=0.5cm]
    \item We first split all triples by relations and form 79 json files for two major paid annotators, each annotator is responsible for around 40 relations.
    \item Annotators should check one relation file at a time for annotating consistency, and start the next one after the former one is finished.
    \item For each relation, annotators are given: (a) The original sentences of the relation with markers indicating the head NP, tail NP and RP extracted by OpenIE. (b) The name (e.g., \emph{composer}), and the Wikidata ID (e.g., \emph{P86}) of the gold relation.
    \item Annotators should first understand the relation by querying the Wikidata, take the relation \emph{``composer (P86)''} as an example, annotators should first query Wikidata through the link  \url{https://www.wikidata.org/wiki/Property:P86} to obtain the definition of the relation and skim through example relational phrases of RPs. The Figure \ref{fig:wikidata_query} shows the Wikidata page containing the definition and examples of \emph{``composer''}.
    \begin{figure}[h]
        \centering
        \scalebox{0.11}{
        \includegraphics{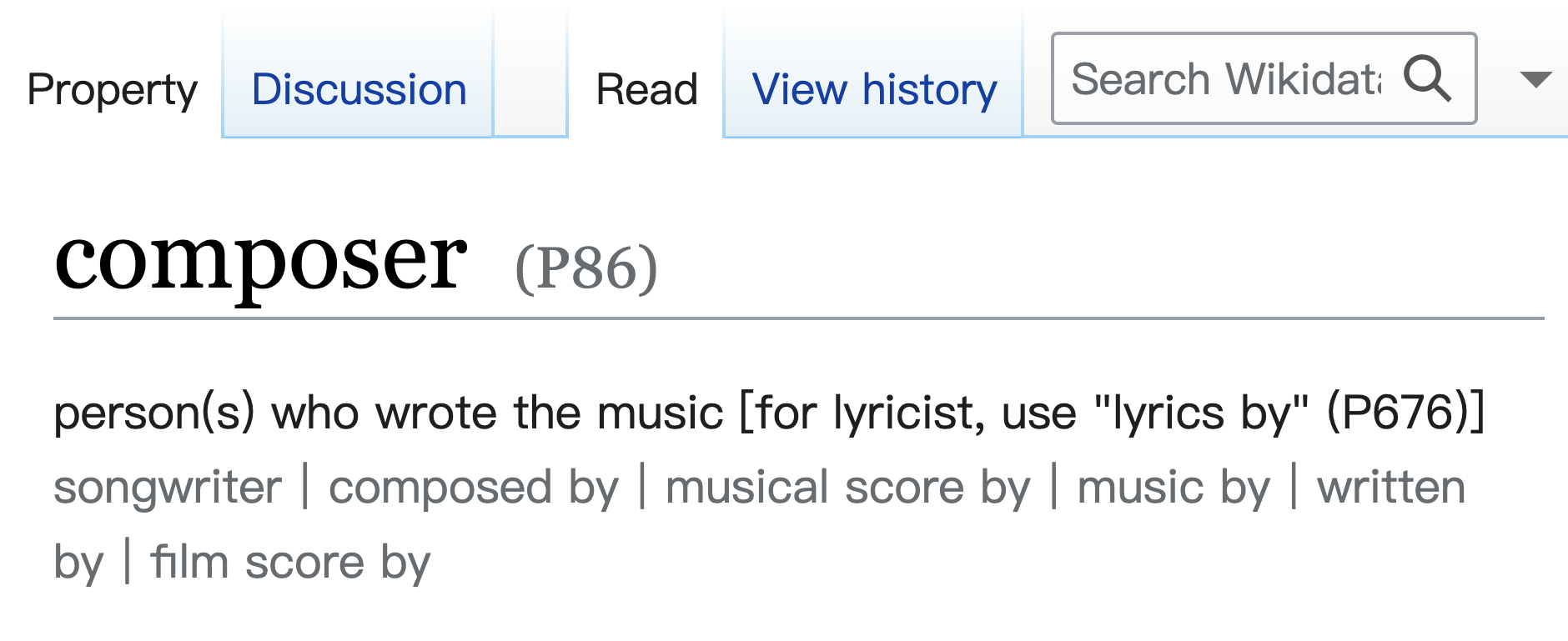}}
        \caption{Wikidata query example.}
        \label{fig:wikidata_query}
    \end{figure}
    \item After fully comprehend the relation, annotators can start to check and revise triples in each file, the details and examples of RP revision and justification of if RP implies the relation are shown in the next two subsection (Appendix \ref{app:detail_guide}, Appendix \ref{app:detail_guide2}).
    \item After two annotators finished their part, we randomly sample 100 samples from each part and ask the annotator responsible for the other part to check. The annotators reached a consensus for approving 97\% of these samples.
\end{enumerate}

\subsection{Guideline for checking the validity of RPs and revision}
\textbf{Definition of relational phrases\ } Relational phrases are textual representations of relations between named entities \cite{relly}, we follow ReVerb \cite{ReVerb2011} and require the relational phrases be continuous span in the sentence. We summarize most cases of relational phrases occurring in our dataset, and the guideline for annotating each case. The annotated RPs are shown in pink.
\begin{itemize}[leftmargin=0.5cm]
    \item \textbf{Case 1: Verb \ } example: \emph{[A] {\color{magenta} married}  [B]}, include different tenses of verbs.
    \item \textbf{Case 2: Verb+preposition\ } example: \\ \emph{[A] {\color{magenta} located at}  [B]}.
    \item \textbf{Case 3: Passive voice \ } example: \emph{[A] {\color{magenta} is designed by} [B]}, \emph{[A] {\color{magenta} is headquartered in} [B]} the linking verb is sometimes omitted: \emph{[A], {\color{magenta} designed by}  [B]}.
    \item \textbf{Case 4: When the tail entity is the appositive of the head entity\ } example: \emph{[A] {\color{magenta}'s son is} [B]}, \emph{[A] {\color{magenta}'s son ,} [B]}, \emph{[A] {\color{magenta}'s masterpiece ,} [B]}, the content between the appositives is usually informative and regarded as RPs.
    \item \textbf{Case 5: Compound predicate} A common case is that the tail NP is a part of the compound predicate of the head NP, e.g., \emph{[A] is the {\color{magenta} civil branch of} [B]}, \emph{[A] is an {\color{magenta} agency of} [B]}, \emph{[A] is a {\color{magenta} novel by} [B]}. When encountering these cases, we do not include the linking verb and article because they are not informative (e.g., \emph{``is the''}).
    \item \textbf{Case 6: Cases omit verb\ } Some cases omit verb, we only annotate the preposition. For example, \emph{[yokosuka arts theatre], part of the bay square complex {\color{magenta}by} [kenzou tange]}, this sentence omits the verb \emph{``built''}.
\end{itemize}
\label{app:detail_guide}

We also provide several revision examples of wrong OpenIE triples, part of them are shown in the Table \ref{tab:bad_openie1} below.
\begin{table*}[h]
\centering
\begin{tabular}{c|c} \toprule
Bad OpenIE RP & \emph{... [the generalitat de catalunya], the governing  \textbf{{\color{teal} body of}} [catalonia], approved ...} \\
Revised      &  \emph{... [the generalitat de catalunya], the \textbf{{\color{magenta}  governing body of}} [catalonia], approved ...} \\ \midrule
Bad OpenIE RP &  \emph{... the ``[althing]'' , the ruling legislative \textbf{{\color{teal}body of}} [iceland]}\\
Revised      &  \emph{... the ``[althing]'' , the  \textbf{{\color{magenta}ruling legislative body of}} [iceland]}\\ \midrule
Bad OpenIE RP &  \emph{... [t-d center] dominion \textbf{{\color{teal} centre}}, designed by [ludwig mies van der rohe] ... } \\
Revised      &  \emph{... [t-d center] dominion centre, \textbf{{\color{magenta} designed by }} [ludwig mies van der rohe] ... } \\ \midrule
Bad OpenIE RP & \thead{\emph{[ace attorney investigations 2] ... and features \textbf{{\color{teal}character designs by}}} \\
                   \emph{tatsuro iwamoto and music by [noriyuki iwadare]}} \\
Revised      &  \thead{\emph{[ace attorney investigations 2] ... and features character designs by} \\
                   \emph{tatsuro iwamoto and \textbf{{\color{magenta} music by }} [noriyuki iwadare]}} \\ \midrule
Bad OpenIE RP             &   \emph{[A] is a \textbf{{\color{teal} farcical musical comedy with}} music by [walter alfred slaughter]}       \\
Revised             & \emph{[A] is a farcical musical comedy with \textbf{{\color{magenta} music by}} [walter alfred slaughter]}  \\ \midrule
Bad OpenIE RP &\emph{[beta cygni a] \textbf{{\color{teal} is a bright}} star from the constellation [cygnus]} \\
Revised & \emph{[beta cygni a] is a \textbf{{\color{magenta} bright star from}} the constellation [cygnus]} \\ \midrule
Bad OpenIE RP & \emph{... [daya district], taichung, \textbf{{\color{teal}taiwan in}} the [chinese taipei]} \\
Revised  & \emph{... [daya district], taichung, taiwan \textbf{{\color{magenta}in}} the [chinese taipei]} \\ \bottomrule
\end{tabular}
\caption{Examples of revising RPs in OpenIE triples.}
\label{tab:bad_openie1}
\end{table*}

\subsection{Guideline for justifying if RP implies the given relation}
\label{app:detail_guide2}
As we stated in the fourth step of the overall annotating process in the Appendix \ref{app:overall1}, we require annotators to fully understand the meaning of the given relation. For each triple, annotators should ask themselves if the relational phrase could express the relation of the head and tail NP in the given context sentence. Note that we don't require the relation could be solely implied by the RP, for example, given the triple and its context: \emph{``[mount elbert] \textbf{{\color{magenta} in}} the [sawatch range] is the highest summit of the rocky mountains''}, it is impossible to infer the relation by the RP \emph{``in''}, but RP is a reasonable text representation of the relation \emph{mountain range} in this context. We found that the extracted RPs can imply the relation in most cases, we show some concrete bad cases to the annotators to help them identify the bad RPs, part of the examples are shown in the Table \ref{tab:bad_openie2}. 
\begin{table*}[h]
\centering
\begin{tabular}{c|c} \toprule
\thead{\emph{\textbf{P1001}} \\ \emph{applies to jurisdiction}} & \thead{\emph{...the process to amend the constitution cannot be initiated in times of war} \\
                                                    \emph{or when the [belgian federal parliament] \textbf{{\color{magenta} is unable to freely meet in}} [belgium]}} \\ \midrule
\thead{\emph{\textbf{P84}} \\ \emph{architect}}     &  {\small \emph{[u. b. city], the \textbf{{\color{magenta} headquarters of}} the  [united breweries group], is a high-end commercial zone. }} \\ \midrule
\thead{\emph{\textbf{P40}} \\ \emph{child}}   & {\small \emph{he was a great-grandson of [berge sigval natanael bergeson], \textbf{{\color{magenta}grand-naphew of}} [ole bergeson]}} \\ \midrule
\thead{\emph{\textbf{P25}} \\ \emph{mother}}   & {\small \emph{bart and [lisa], \textbf{{\color{magenta} sent out of}} the house by [marge simpson] ... }} 
\\ \bottomrule
\end{tabular}
\caption{Examples of bad RPs that cannot imply the given relations.}
\label{tab:bad_openie2}
\end{table*}

\section{Classic Metrics}
Gold cluster assignment: $\mathcal{G} = \{C^g_1 \dotsc C^g_{K}\}$, predicted cluster assignment: $\mathcal{P}=\{C^p_1 \dotsc C^p_M\}$, where $C^g_i$ and $C_j^p$ are gold and predicted cluster respectively.
\paragraph{Macro Metrics}
\begin{equation}
    \begin{aligned}
    P_{macro}(\mathcal{G}, \mathcal{P}) = \frac{\vert \{p \in \mathcal{P} \vert \exists g \in \mathcal{G}, p \subseteq g \}\vert}{ \vert \mathcal{P} \vert }  \\
    R_{macro}(\mathcal{G}, \mathcal{P}) = P_{micro}(\mathcal{P}, \mathcal{G})
    \end{aligned}
\end{equation}

\paragraph{Micro Metrics}
\begin{equation}
    \begin{aligned}
    P_{micro}(\mathcal{G}, \mathcal{P}) = \frac{\sum_{g \in \mathcal{G}} \max_{p \in \mathcal{P}}\vert g \cap p \vert}{N}  \\
    R_{micro}(\mathcal{G}, \mathcal{P}) = P_{micro}(\mathcal{P}, \mathcal{G})
    \end{aligned}
\end{equation}
Where $N$ is the total number of different phrases that appear in $\mathcal{G}$ (or $\mathcal{P}$).

\paragraph{Pairwise Metrics}
\begin{equation}
    \begin{aligned}
    P_{pair}(\mathcal{G}, \mathcal{P}) =\\ \frac{\sum_{p \in \mathcal{P}}\vert \{ (u, u^\prime) \in p, \exists g\in \mathcal{G}, \forall (u, u^\prime) \in p \} \vert}{\sum_{p \in \mathcal{P}} C^{|p|}_2}  \\
    R_{pair}(\mathcal{G}, \mathcal{P}) =\\ \frac{\sum_{p \in \mathcal{P}}\vert \{ (u, u^\prime) \in p, \exists g\in \mathcal{G}, \forall (u, u^\prime) \in p \} \vert}{\sum_{g \in \mathcal{G}} C^{|g|}_2}
    \end{aligned}
\end{equation}
For more details about classic metrics, please refer to the Sec. 7.2 of the CESI paper \cite{vashishth2018cesi}.

\section{Extension of Micro Metrics}
\label{app:mi}
Gold overlapping clusters: $\textit{NPC-O} = \{C^O_1 \dotsc C^O_{K_3}\}$, predicted clusters $\mathcal{P}=\{C^p_1 \dotsc C^p_M\}$.
\begin{equation}
    \begin{aligned}
    P_{micro}(\textit{NPC-O}, \mathcal{P}) = \frac{\sum_{g \in \textit{NPC-O}} \max_{p \in \mathcal{P}}\vert g \cap p \vert}{\sum_{g \in \textit{NPC-O}} \vert g \vert }  \\
    R_{micro}(\textit{NPC-O}, \mathcal{P}) = P_{micro}(\mathcal{P}, \textit{NPC-O})
    \end{aligned}
\end{equation}
We modify the denominator compared to the micro metric in \cite{vashishth2018cesi}.

\section{Standardization}
\label{app:std}
Following \citet{timkey2021all}, we perform standardization for phrase embeddings to remove the rogue dimensions. Denote $\bm{E}_\mathcal{R} \in \mathbb{R}^{N\times D}$ as the embedding matrix of all RP phrases, where $N$ is the number of triples, and $D$ is the dimension of contextual embedding. The standardized RP embedding matrix $\bm{E}_\mathcal{R}^\prime$ is:
\begin{equation}
    \begin{aligned}
        \bm{\mu} &= \frac1N \sum_i^N \bm{E}_\mathcal{R}[i]\\
        \bm{\sigma} &= \sqrt{\frac1N \sum_i^N (\bm{E}_\mathcal{R}[i]-\bm{\mu})^2} \\
        \bm{E}_\mathcal{R}^\prime[i] &= \frac{\bm{E}_\mathcal{R}[i] - \bm{\mu}}{\bm{\sigma}}
    \end{aligned}
\end{equation}
We empirically find that the standardized phrase embedding is better than the original one in most cases.

\section{Full Result}
\label{app:fullres}
We show the full results in Table \ref{tab:fullres1} and Table \ref{tab:fullres2}.
\begin{table*}[h]
\centering
\scalebox{0.85}{
\begin{tabular}{l|llll|llll|lll} \toprule
                              & \multicolumn{4}{c}{\textit{\textbf{NPC-E (subject)}}}                                        & \multicolumn{4}{c}{\textit{\textbf{NPC-E (object)}}}                                         & \multicolumn{3}{c}{\textit{\textbf{RPC}}}                             \\
\multirow{-2}{*}{}            & \textit{\textbf{Ma}} & \textit{\textbf{Mi}} & \textit{\textbf{Pair}} & \textit{\textbf{AVG}} & \textit{\textbf{Ma}} & \textit{\textbf{Mi}} & \textit{\textbf{Pair}} & \textit{\textbf{AVG}} & \textit{\textbf{Mi}} & \textit{\textbf{Pair}} & \textit{\textbf{AVG}} \\ \midrule
\textit{Token Sim+SI}         & 86.85                & 88.41                & 73.43                  & 82.90                 & 78.43                & 85.34                & 74.29                  & 79.35                 & 50.26                & 17.62                  & 33.94                 \\
\textit{Random+HAC}           & 87.97                & 89.90                & 78.10                  & 85.32                 & 81.24                & 88.11                & 85.97                  & 85.11                 & 48.47            & 23.49                  & 35.98                 \\
\textit{GloVe+HAC}            & 85.63                & 87.22                & 62.07                  & 78.31                 & 79.90                & 87.06                & 92.74                  & 86.57                 & 47.92             & 23.22                 & 35.57                 \\
\textit{GloVe+HolE+HAC}       & 88.74                & 89.59                & 62.06                  & 80.13                 & 80.85                & 88.20                & 92.95                  & 87.33                 & 27.05                & 8.80                   & 17.93                 \\
\textit{GloVe+SI+HAC}         & 88.72                & 89.82                & 62.72                  & 80.42                 & 80.85                & 88.21                & 92.95                  & 87.34                 & 27.02                & 8.79                   & 17.91                 \\
\textit{CESI}                 & 88.72                & 89.58                & 62.02                  & 80.11                 & 82.93                & 88.37                & 89.17                  & 86.82                 & 27.31                & 9.43                   & 18.37                 \\
\textit{VAEGMM+SI}            & 85.63                & 87.51                & 69.44                  & 80.86                 & 78.55                & 85.93                & 84.26                  & 82.91                 & 47.86                & 20.34                  & 34.10                 \\
\textit{VAEGMM+HolE}          & 85.50                & 87.26                & 67.69                  & 80.15                 & 78.54                & 85.91                & 84.17                  & 82.87                 & 47.92                & 24.31                  & 36.12                 \\
\textit{CUVA}                 & 85.58                & 87.45                & 69.00                  & 80.68                 & 78.56                & 85.95                & 84.33                  & 82.95                 & 47.94                & 24.32                  & 36.13   \\ \hline \hline
\rowcolor[HTML]{ECF4FF} 
\textit{Bert-base}            & 90.84                & 92.11                & 77.84                  & 86.93                 & 86.84                & 90.53                & 83.36                  & 86.91                 & 55.85 &	53.09 &	54.47              \\
\rowcolor[HTML]{ECF4FF} 
\textit{Roberta-base}         & 87.74                & 89.59                & 71.22                  & 82.85                 & 83.12                & 88.22                & 83.66                  & 85.00                 & 44.81 &	37.35 & 41.08                \\
\rowcolor[HTML]{ECF4FF} 
\textit{SpanBert-base}        & 91.14                & 91.86                & 70.14                  & 84.38                 & 85.82                & 89.92                & 83.22                  & 86.32                 & 43.49 &	39.36 & 44.04               \\
\rowcolor[HTML]{ECF4FF} 
\textit{ERNIE2.0-base}        & 91.19                & 92.50                & 76.35                  & 86.68                 & 83.15                & 88.89                & 92.29                  & 88.11                 &53.94 &	55.38 &	54.66                \\ \hline
\rowcolor[HTML]{ECF4FF} 
\textit{Bert-base-triple}     & 90.05                & 91.57                & 76.41                  & 86.01                 & 83.57                & 89.67                & 93.51                  & 88.92                 & 57.53 &	59.36 &	58.45              \\
\rowcolor[HTML]{ECF4FF} 
\textit{Roberta-base-triple}  & 90.40                & 91.05                & 74.66                  & 85.37                 & 81.07                & 87.96                & 92.62                  & 87.22                 & 52.75 &	48.86 &	50.81             \\
\rowcolor[HTML]{ECF4FF} 
\textit{SpanBert-base-triple} & 91.95                & 91.87                & 73.36                  & 85.73                 & 85.26                & 89.19                & 83.23                  & 85.89                 & 48.09 &	44.26 &	46.18          \\
\rowcolor[HTML]{ECF4FF} 
\textit{ERNIE2.0-base-triple} & 91.14                & 92.34                & 78.16                  & 87.21                 & 85.71                & 90.12                & 84.96                  & 86.93                 & 56.58 &	57.98 &	57.28           \\ \hline
\rowcolor[HTML]{ECF4FF} 
\textit{Bert-large}           & 90.47                & 91.93                & 78.87                  & 87.09                 & 83.81                & 89.78                & 93.56                  & 89.05                 & 53.81 &	46.81 & 50.31                \\
\rowcolor[HTML]{ECF4FF} 
\textit{Roberta-large}        & 88.34                & 90.14                & 72.03                  & 83.50                 & 84.48                & 89.01                & 83.93                  & 85.81                 & 43.01 & 38.02 & 40.51               \\
\rowcolor[HTML]{ECF4FF} 
\textit{SpanBert-large}       & 85.82                & 89.92                & 83.22                  & 86.32                 & 87.14                & 90.81                & 82.07                  & 86.67                 & 45.5 & 46.18. & 45.84                 \\
\rowcolor[HTML]{ECF4FF} 
\textit{ERNIE2.0-large}       & 91.06                & 92.26                & 75.30                  & 86.21                 & 83.09                & 89.56                & 93.92                  & 88.86                 & 49.48 &	48.32 & 48.90                \\ 
\hline
\rowcolor[HTML]{ECF4FF} 
\textit{Bert-large-triple}    & 90.02 & 92.35 & 93.64 & 92.00 & 87.47 & 90.99 & 84.49 & 87.65 & 56.27 & 60.04 & 58.16  \\ 
\rowcolor[HTML]{ECF4FF} 
\textit{Roberta-large-triple} & 87.93 & 89.35 & 72.93 & 83.40 & 84.24 & 88.57 & 83.56 & 85.46 & 47.37 & 41.41 & 44.39  \\ 
\rowcolor[HTML]{ECF4FF} 
\textit{SpanBert-large-triple} & 90.57 & 94.26 & 94.74 & 93.19 & 85.87 & 89.80 & 82.34 & 86.00 & 49.24 & 45.68 & 47.46  \\ 
\rowcolor[HTML]{ECF4FF} 
\textit{ERNIE2.0-large-triple} & 93.32 & 94.41 & 85.90 & 91.21 & 86.19 & 90.65 & 91.21 & 89.35 & 54.37 & 53.32 & 53.85  \\
\bottomrule
\end{tabular}}
\caption{Full results of NPC-E and RPC.}
\label{tab:fullres1}
\end{table*}

\begin{table*}[]
\centering
\scalebox{0.66}{
\begin{tabular}{l|llll|lll|llll|lll} \toprule
                              & \multicolumn{7}{c|}{\textit{\textbf{NPC-O (subject)}}}                                                                                                                          & \multicolumn{7}{c}{\textit{\textbf{NPC-O (object)}}}                                                                                                                           \\
\multirow{-2}{*}{}            & \textit{\textbf{Ma}} & \textit{\textbf{Mi}} & \textit{\textbf{Pair}} & \textit{\textbf{Avg}} & $J_{g \to p}$ & $J_{p \to g}$ & \textit{\textbf{Avg}} & \textit{\textbf{Ma}} & \textit{\textbf{Mi}} & \textit{\textbf{Pair}} & \textit{\textbf{Avg}} & $J_{g\to p}$ & $J_{p \to g}$ & \textit{\textbf{Avg}} \\ \midrule
\textit{Token Sim+SI}                  & 63.28                & 35.86                & 0.29                   & 33.14                 & -                          &  -                         & -                       & 66.24                & 51.61                & 3.91                   & 40.59                 &    -                        &  -                          &  -                     \\
\textit{Random+HAC}                    & 69.04                & 42.20                & 0.69                   & 37.31                 & 62.54                      & 15.05                      & 38.79                 & 74.72                & 54.25                & 5.72                   & 44.90                 & 66.78                      & 22.02                      & 44.40                 \\
\textit{GloVe+HAC}                     & 70.37                & 42.80                & 0.84                   & 38.00                 & 63.00                      & 15.07                      & 39.04                 & 74.08                & 60.25                & 7.73                   & 47.35                 & 68.73                      & 22.17                      & 45.45                 \\
\textit{GloVe+HolE+HAC}                & 71.41                & 46.06                & 1.43                   & 39.63                 & 63.95                      & 15.78                      & 39.87                 & 76.68                & 60.38                & 7.68                   & 48.25                 & 71.06                      & 23.30                      & 47.18                 \\
\textit{GloVe+SI+HAC}                  & 71.68                & 46.32                & 1.62                   & 39.87                 & 63.88                      & 15.78                      & 39.83                 & 76.63                & 60.43                & 7.72                   & 48.26                 & 70.96                       & 23.26                      & 47.11              \\
\textit{CESI}                          & 71.66                & 46.51                & 1.62                   & 39.93                 & 63.98                      & 15.90                      & 39.89                 & 76.57                & 60.45                & 7.74                   & 48.25                 & 71.14                       & 23.24                       & 47.19              \\
\textit{VAEGMM+SI}                     & 69.32                & 42.08                & 0.83                   & 37.41                 & -                          & -                          & -                      & 72.52                & 60.10                & 8.07                   & 46.90                 &    -                        &   -                         &   -                    \\
\textit{VAEGMM+HolE}                   & 69.09                & 41.81                & 0.75                   & 37.22                 & -                          & -                          & -                      & 72.48                & 60.09                & 8.07                   & 46.88                 &   -                         &   -                         &   -                    \\
\textit{CUVA}                          & 68.76                & 41.76                & 0.76                   & 37.09                 & -                          & -                          & -                       & 72.48                & 60.12                & 8.08                   & 46.89                 &    -                        &   -                         &         -              \\ \hline \hline
\rowcolor[HTML]{ECF4FF} 
\textit{Bert-base}            & 78.77                & 47.20                & 2.93                   & 42.97                 & 73.27                      & 15.04                      & 44.16                 & 68.10                & 63.71                & 20.32                  & 50.71                 & 76.41                      & 17.56                      & 46.99                 \\
\rowcolor[HTML]{ECF4FF} 
\textit{Roberta-base}         & 74.18                & 42.67                & 0.78                   & 39.21                 & 68.06                      & 14.42                      & 41.24                 & 78.22                & 54.17                & 5.81                   & 46.07                 & 71.24                      & 17.04                      & 44.14                 \\
\rowcolor[HTML]{ECF4FF} 
\textit{SpanBert-base}        & 79.00                & 44.58                & 1.25                   & 41.61                 & 71.54                      & 14.78                      & 43.16                 & 80.58                & 55.29                & 6.00                   & 47.29                 & 68.52                      & 22.18                      & 45.35                 \\
\rowcolor[HTML]{ECF4FF} 
\textit{ERNIE2.0-base}        & 78.64                & 46.85                & 2.31                   & 42.60                 & 72.93                      & 15.17                      & 44.05                 & 68.15                & 66.89                & 21.76                  & 52.27                 & 71.81                      & 22.29                      & 47.05                 \\ \hline
\rowcolor[HTML]{ECF4FF} 
\textit{Bert-base-triple}     & 80.71                & 47.58                & 2.84                   & 43.71                 & 75.02                      & 15.36                      & 45.19                 & 67.87                & 65.96                & 21.51                  & 51.78                 & 77.56                      & 17.42                      & 47.49                 \\
\rowcolor[HTML]{ECF4FF} 
\textit{Roberta-base-triple}  & 78.49                & 46.94                & 1.44                   & 42.29                 & 73.36                      & 15.31                      & 44.33                 & 82.31                & 60.14                & 8.47                   & 50.31                 & 76.09                      & 17.53                      & 46.81                 \\
\rowcolor[HTML]{ECF4FF} 
\textit{SpanBert-base-triple} & 80.49                & 45.47                & 1.63                   & 42.53                 & 73.47                      & 14.98                      & 44.23                 & 80.17                & 56.15                & 6.49                   & 47.60                 & 68.84                      & 22.29                      & 45.56                 \\
\rowcolor[HTML]{ECF4FF} 
\textit{ERNIE2.0-base-triple} & 79.10                & 47.61                & 2.91                   & 43.21                 & 73.51                      & 15.15                      & 44.33                 & 69.07                & 66.02                & 16.90                  & 50.66                 & 72.22                      & 22.33                      & 47.27                 \\ \hline
\rowcolor[HTML]{ECF4FF} 
\textit{Bert-large}           & 77.93                & 46.33                & 2.76                   & 42.34                 & 73.09                      & 15.22                      & 44.16                 & 71.92                & 63.47                & 16.75                  & 50.71                 & 76.92                      & 17.58                      & 47.25                 \\
\rowcolor[HTML]{ECF4FF} 
\textit{Roberta-large}        & 75.16                & 43.63                & 0.86                   & 39.88                 & 69.96                      & 14.73                      & 42.35                 & 78.54                & 59.72                & 6.80                   & 48.35                 & 69.07                      & 22.09                      & 45.58                 \\
\rowcolor[HTML]{ECF4FF} 
\textit{SpanBert-large}       & 77.40                & 43.99                & 1.48                   & 40.96                 & 71.08                      & 14.72                      & 42.90                 & 81.05                & 56.38                & 6.33                   & 47.92                 & 69.00                      & 22.36                      & 45.68                 \\
\rowcolor[HTML]{ECF4FF} 
\textit{ERNIE2.0-large}       & 76.86                & 48.16                & 3.11                   & 42.71                 & 72.90                      & 15.11                      & 44.01                 & 71.49                & 64.90                & 19.56                  & 51.98                 & 72.06                      & 22.39                      & 47.22                 \\  \hline
\rowcolor[HTML]{ECF4FF} 
\textit{Bert-large-triple}   & 80.19 & 47.92 & 2.11 & 43.41 & 75.46 & 15.58 & 45.52 & 66.59 & 64.34 & 21.86 & 50.93 & 77.63 & 17.96 & 47.79  \\ 
\rowcolor[HTML]{ECF4FF} 
\textit{Roberta-large-triple}    & 77.56 & 46.91 & 1.99 & 42.15 & 71.82 & 14.99 & 43.40 & 76.83 & 59.18 & 7.64  & 47.88 & 74.44 & 17.42 & 45.93  \\ 
\rowcolor[HTML]{ECF4FF} 
\textit{SpanBert-large-triple}  & 82.09 & 44.88 & 0.62 & 42.53 & 70.47 & 14.71 & 42.59 & 81.97 & 56.08 & 5.81  & 47.95 & 73.79 & 16.85 & 45.32  \\ 
\rowcolor[HTML]{ECF4FF} 
\textit{ERNIE2.0-large-triple}   & 79.81 & 48.04 & 3.04 & 43.63 & 75.28 & 15.52 & 45.40 & 71.23 & 64.80 & 20.93 & 52.32 & 76.67 & 17.71 & 47.19   \\ 
\bottomrule
\end{tabular}}
\caption{Full results of NPC-O.}
\label{tab:fullres2}
\end{table*}

\section{Encoding Strategy Comparison}
\label{app:encode}

\begin{figure}[!h]
    \centering
    \scalebox{0.23}{
    \includegraphics{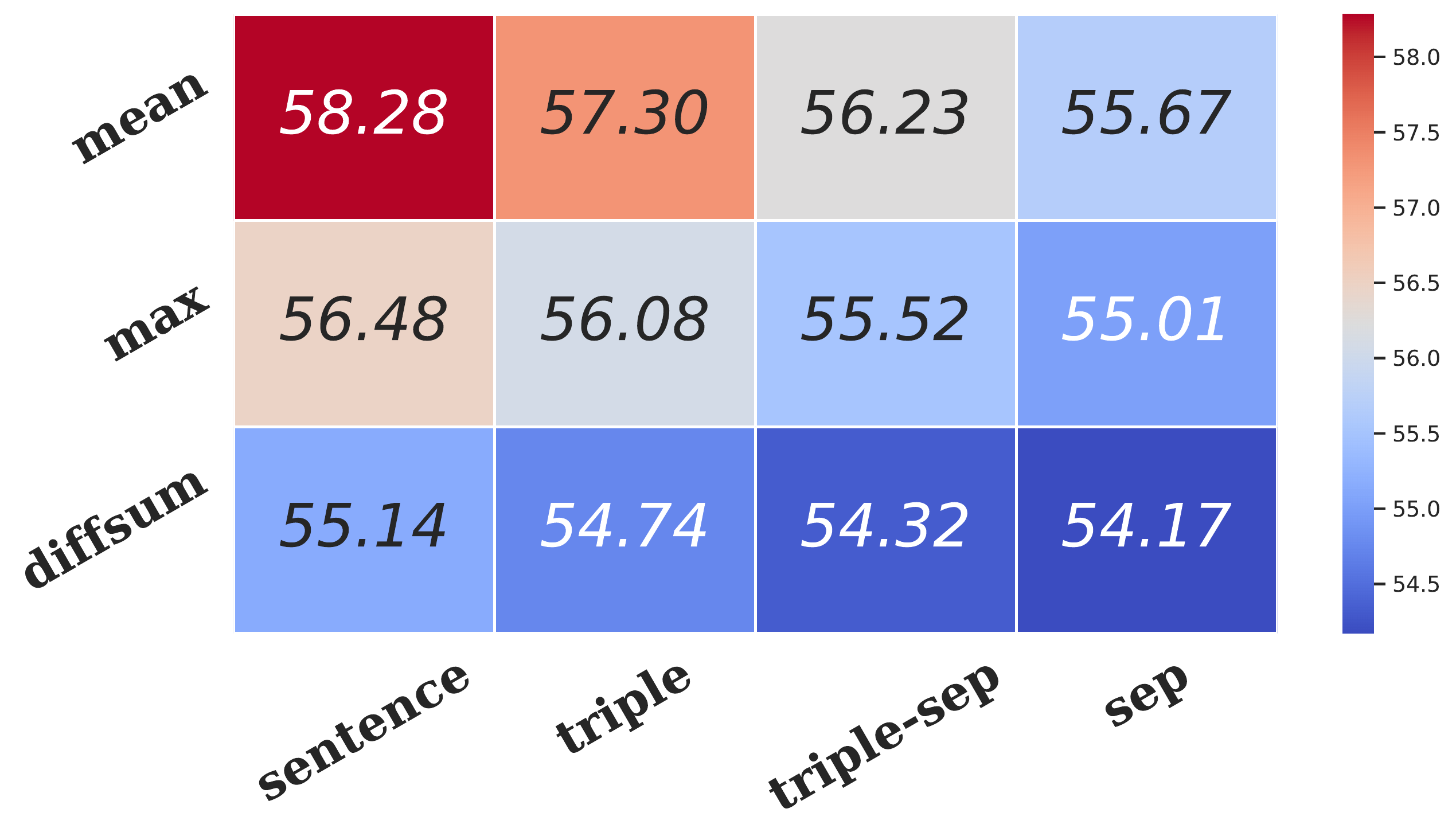}}
    \caption{Overall average performance of different encoding strategies.}
    \label{fig:res_encode}
\end{figure}

\begin{figure*}[h]
     \centering
     \begin{subfigure}[b]{0.32\textwidth}
         \centering
         \includegraphics[width=\textwidth]{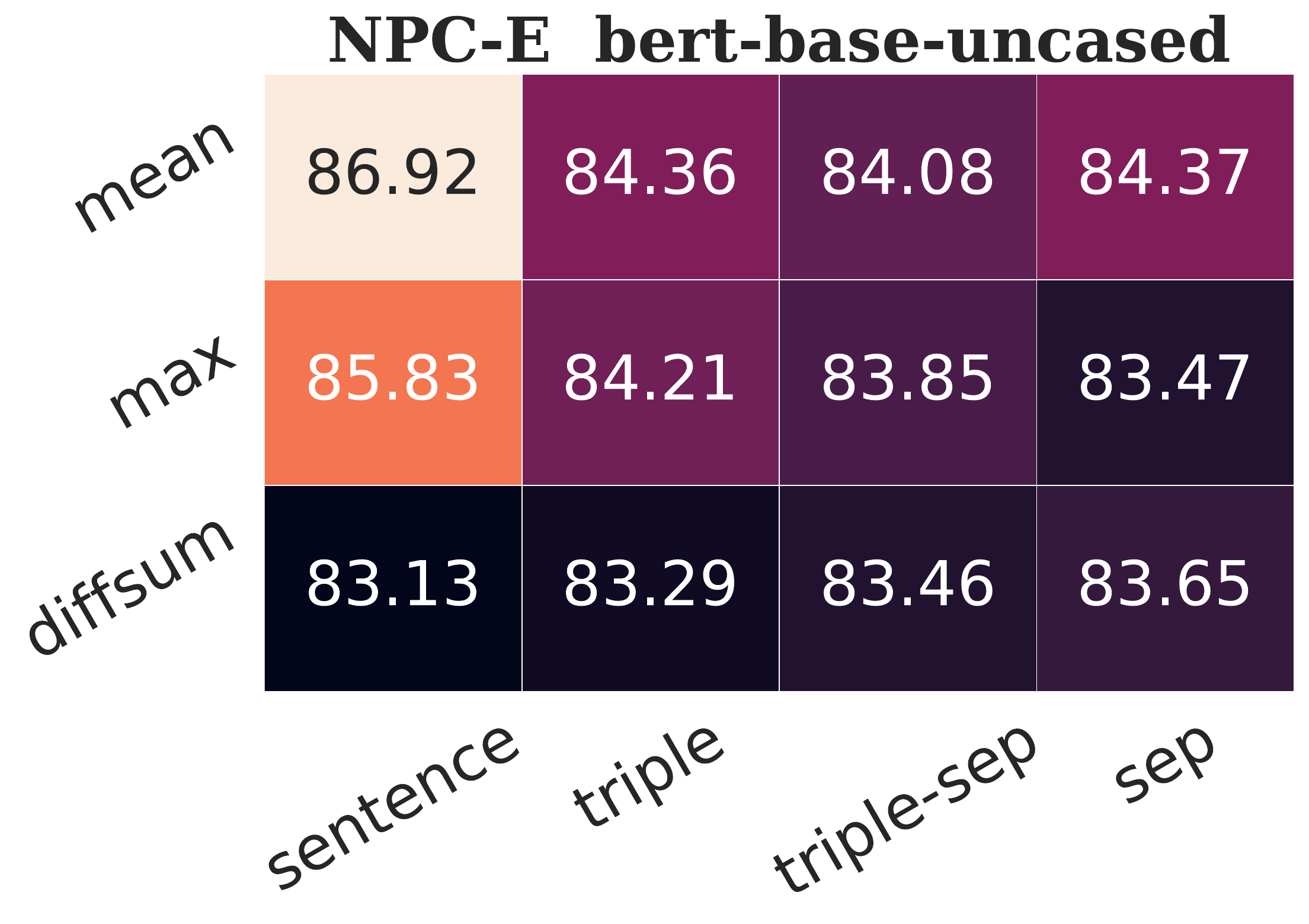}
     \end{subfigure}
     \hfill
     \begin{subfigure}[b]{0.32\textwidth}
         \centering
         \includegraphics[width=\textwidth]{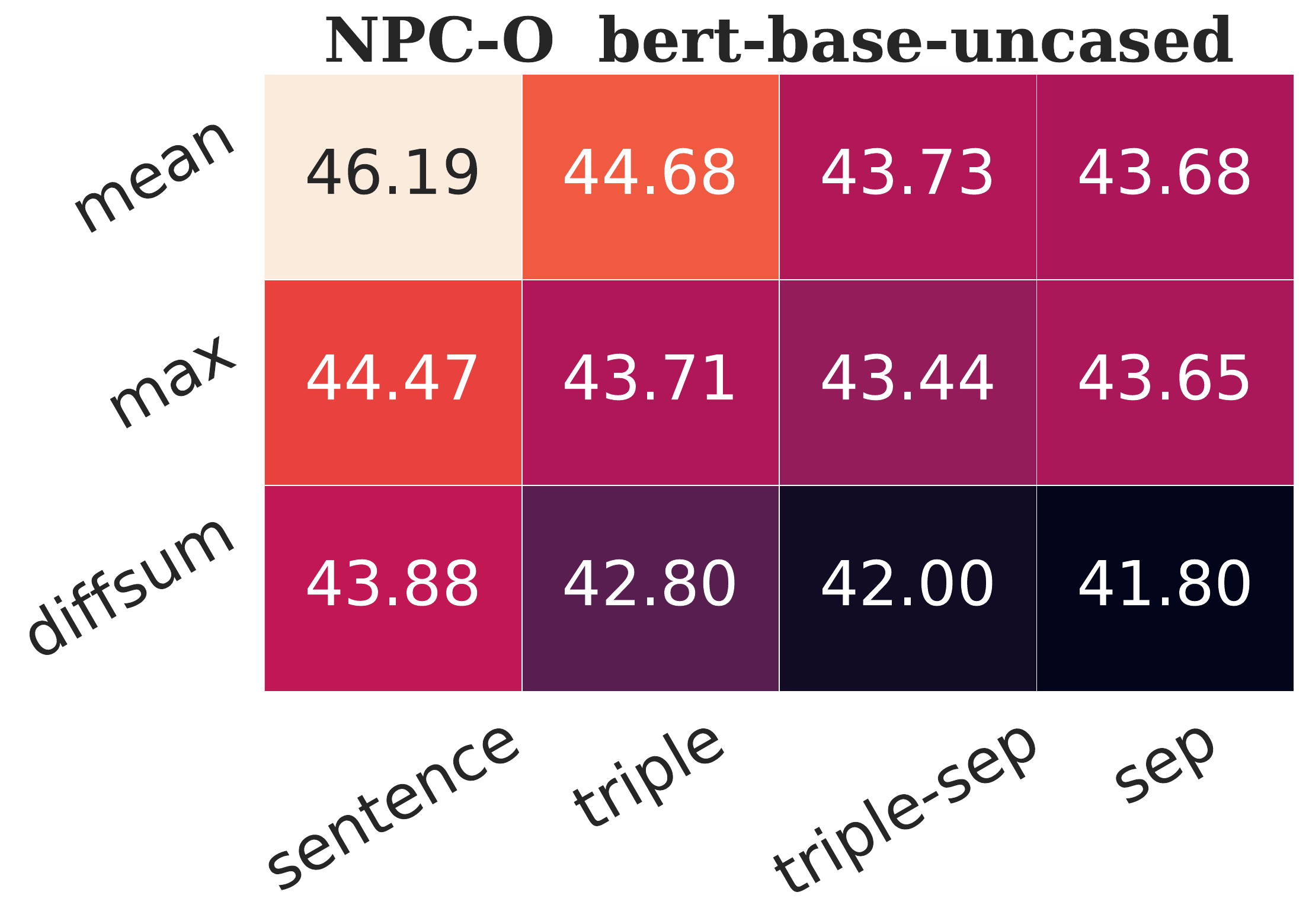}
     \end{subfigure}
     \hfill
     \begin{subfigure}[b]{0.32\textwidth}
         \centering
         \includegraphics[width=\textwidth]{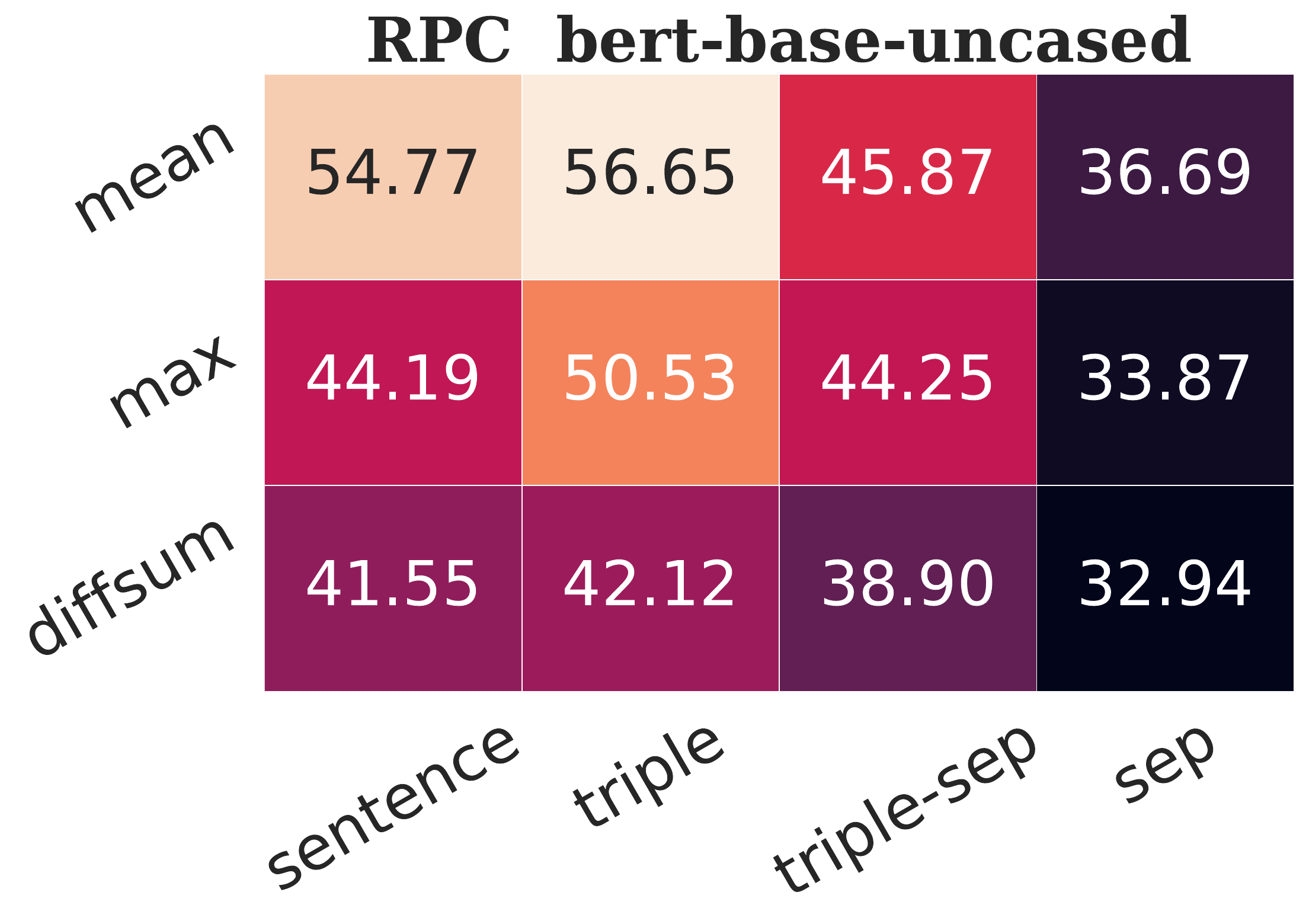}
     \end{subfigure}
     \hfill
     \begin{subfigure}[b]{0.32\textwidth}
         \centering
         \includegraphics[width=\textwidth]{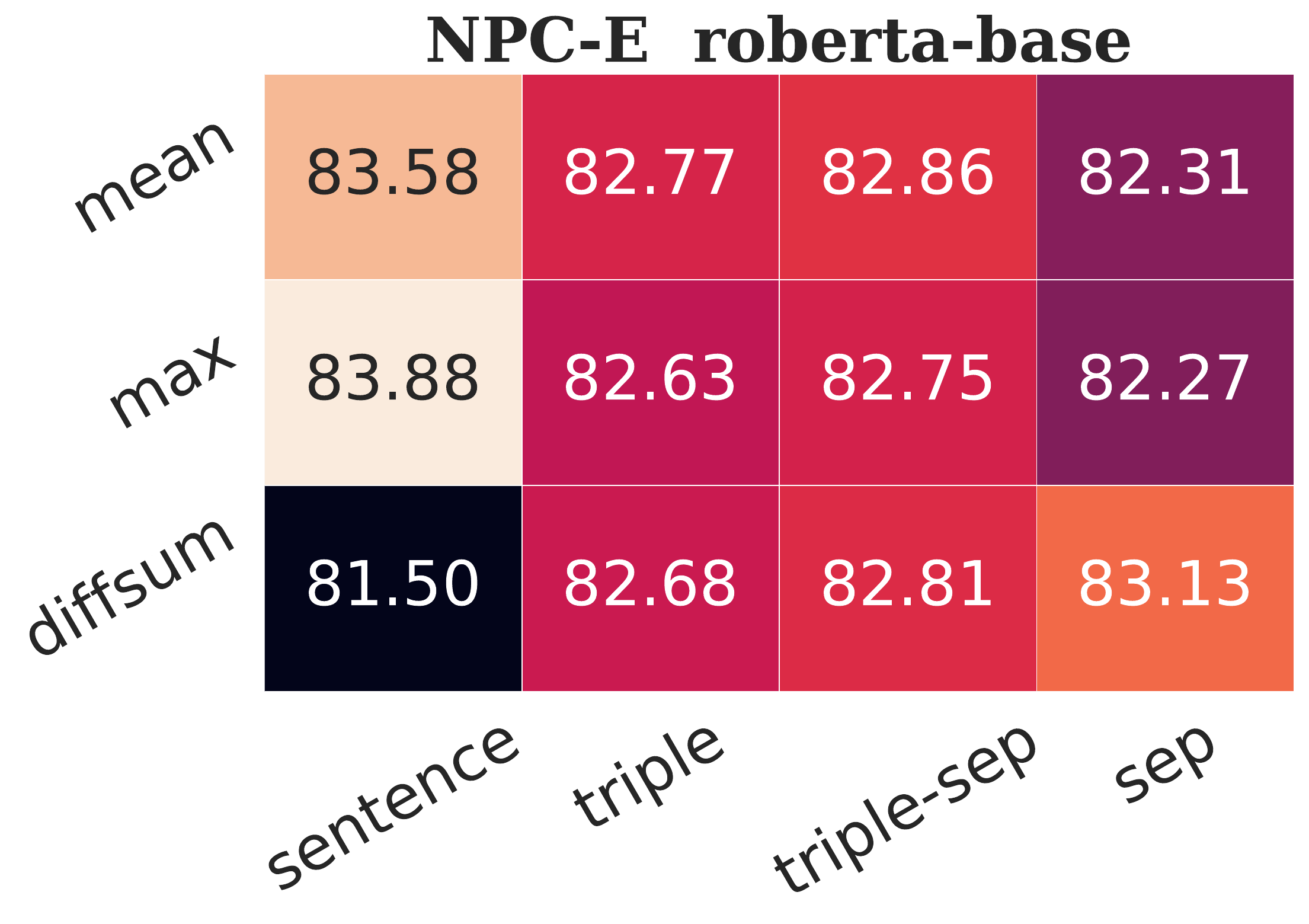}
     \end{subfigure}
     \hfill
     \begin{subfigure}[b]{0.32\textwidth}
         \centering
         \includegraphics[width=\textwidth]{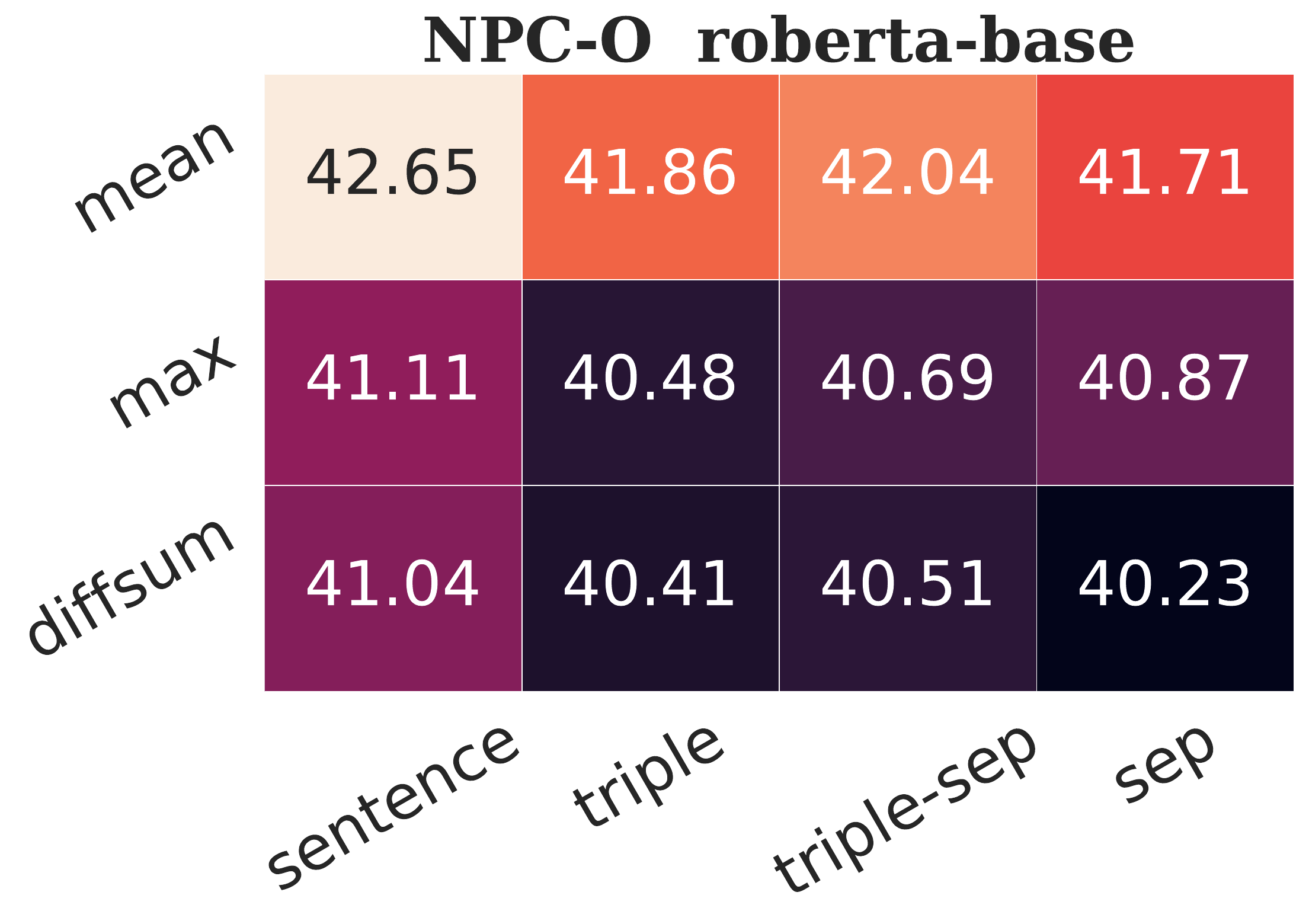}
     \end{subfigure}
     \hfill
     \begin{subfigure}[b]{0.32\textwidth}
         \centering
         \includegraphics[width=\textwidth]{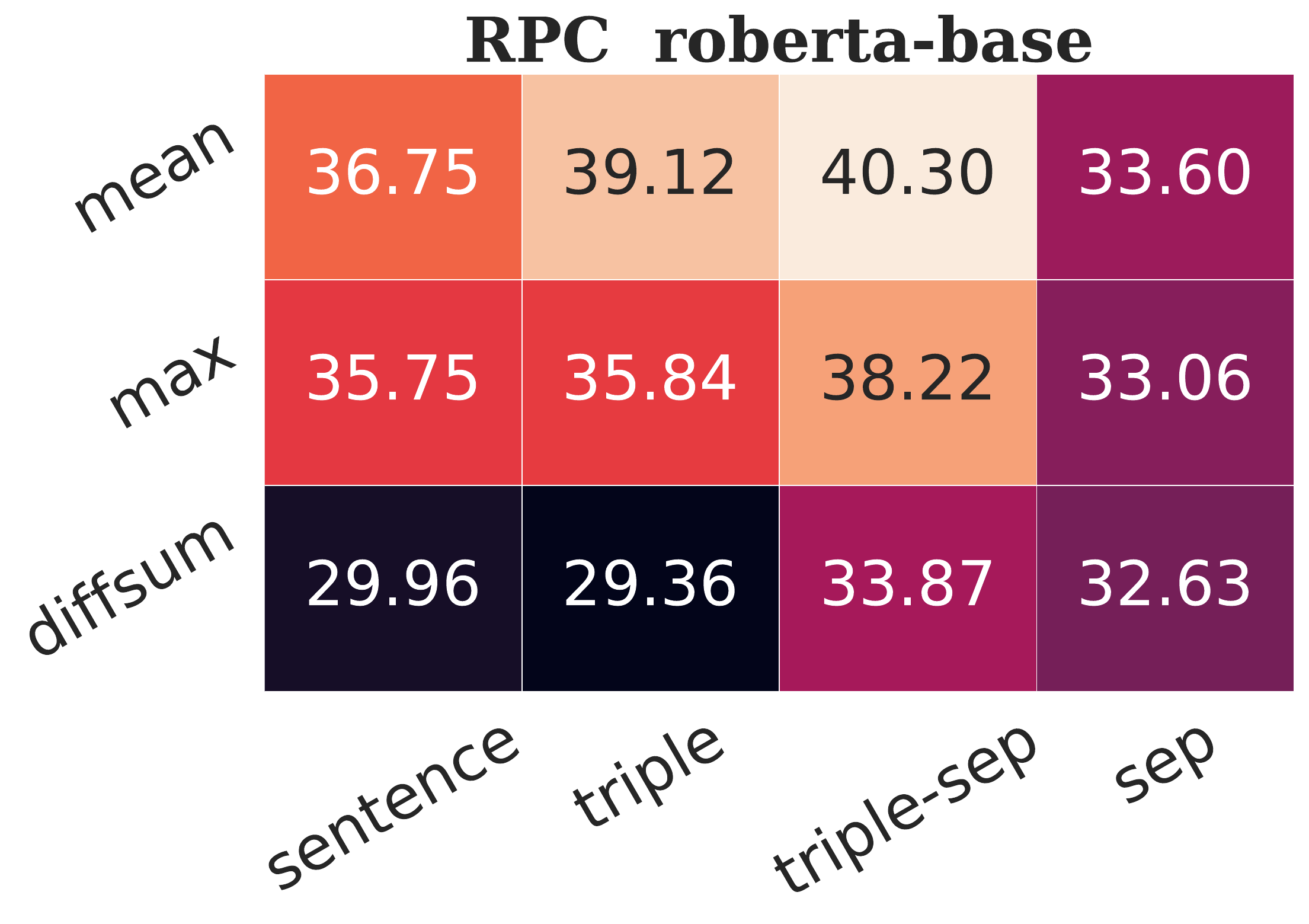}
     \end{subfigure}
     \hfill
     \begin{subfigure}[b]{0.32\textwidth}
         \centering
         \includegraphics[width=\textwidth]{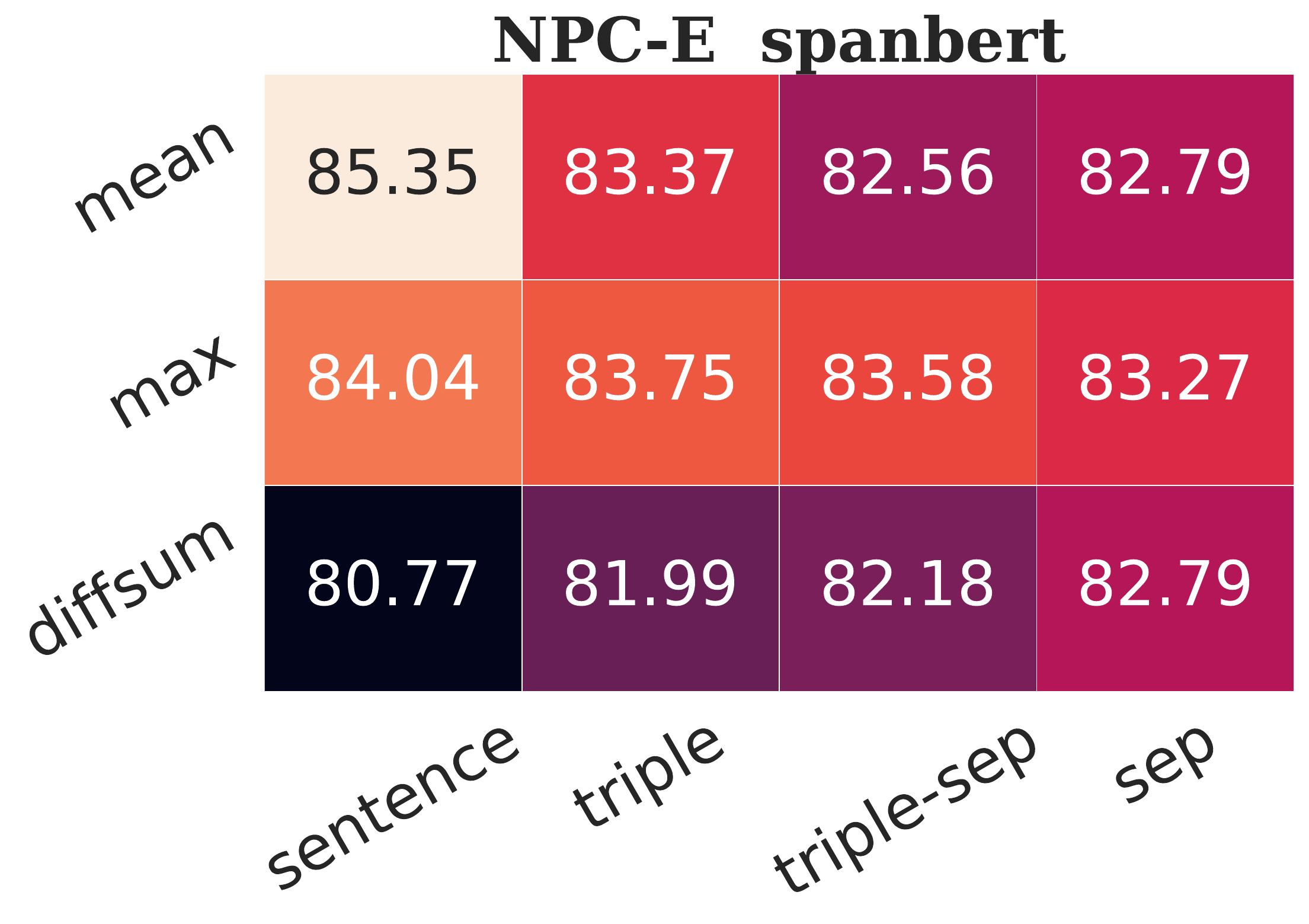}
     \end{subfigure}
     \hfill
     \begin{subfigure}[b]{0.32\textwidth}
         \centering
         \includegraphics[width=\textwidth]{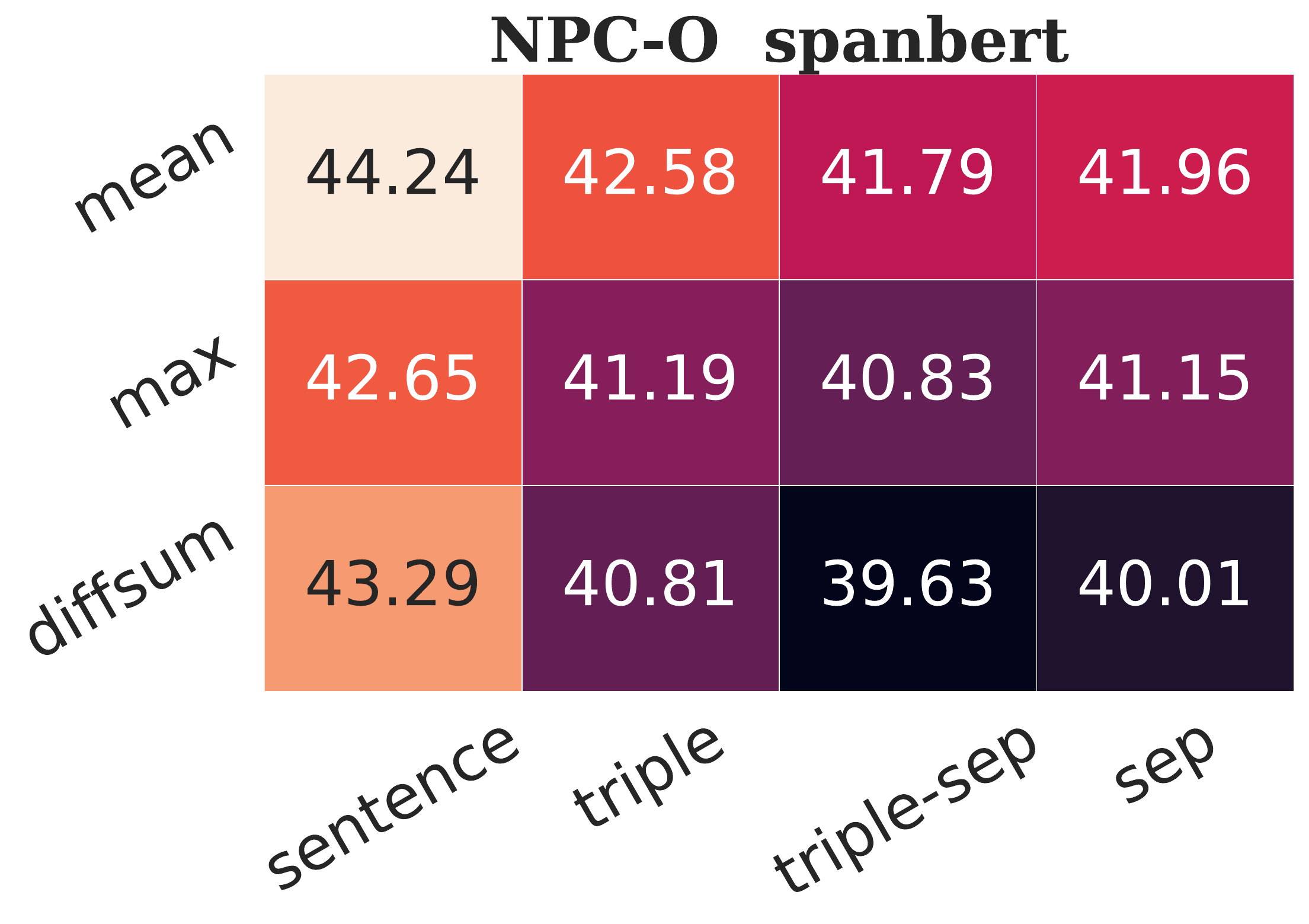}
     \end{subfigure}
     \hfill
     \begin{subfigure}[b]{0.32\textwidth}
         \centering
         \includegraphics[width=\textwidth]{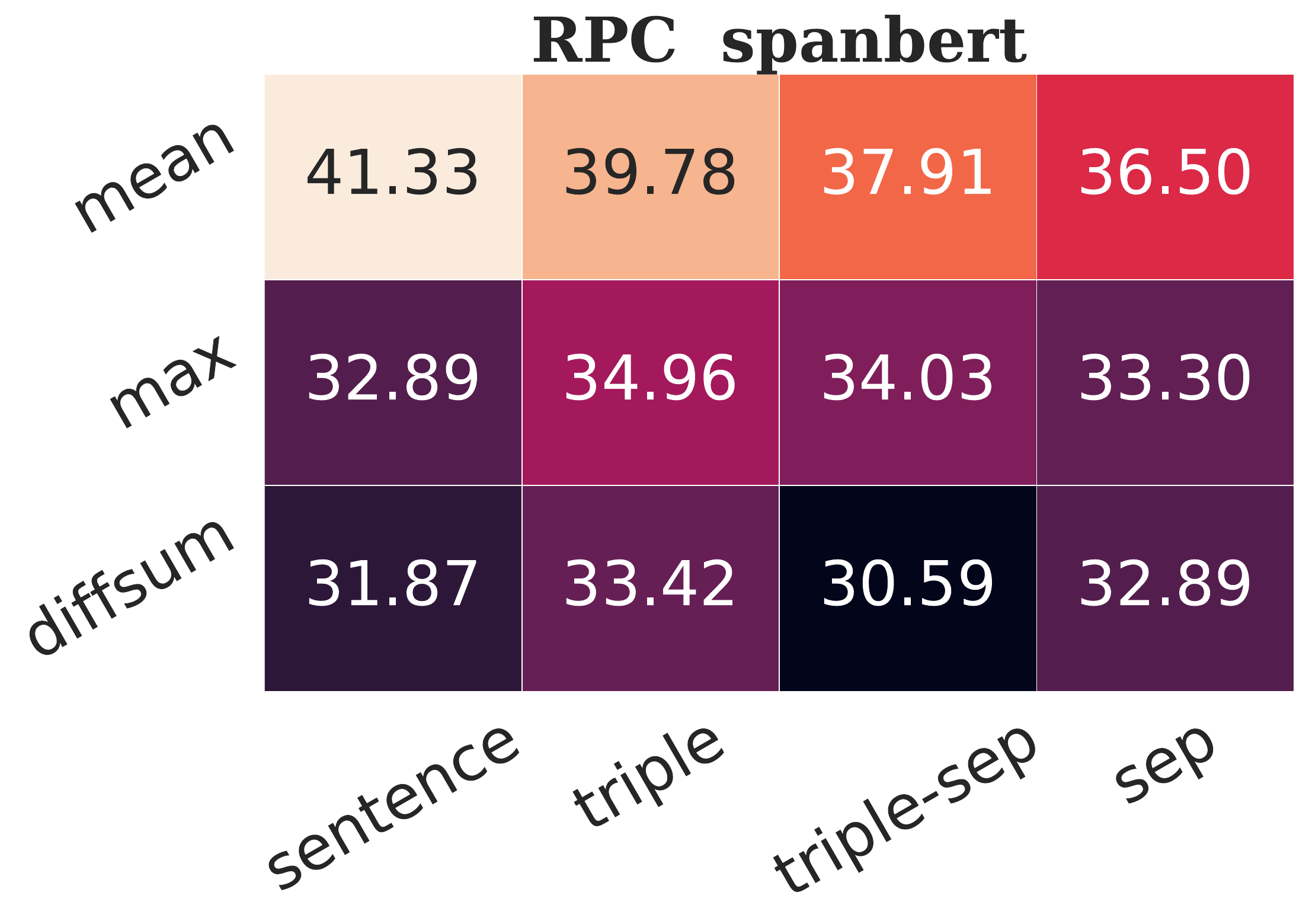}
     \end{subfigure}
     \hfill
          \begin{subfigure}[b]{0.32\textwidth}
         \centering
         \includegraphics[width=\textwidth]{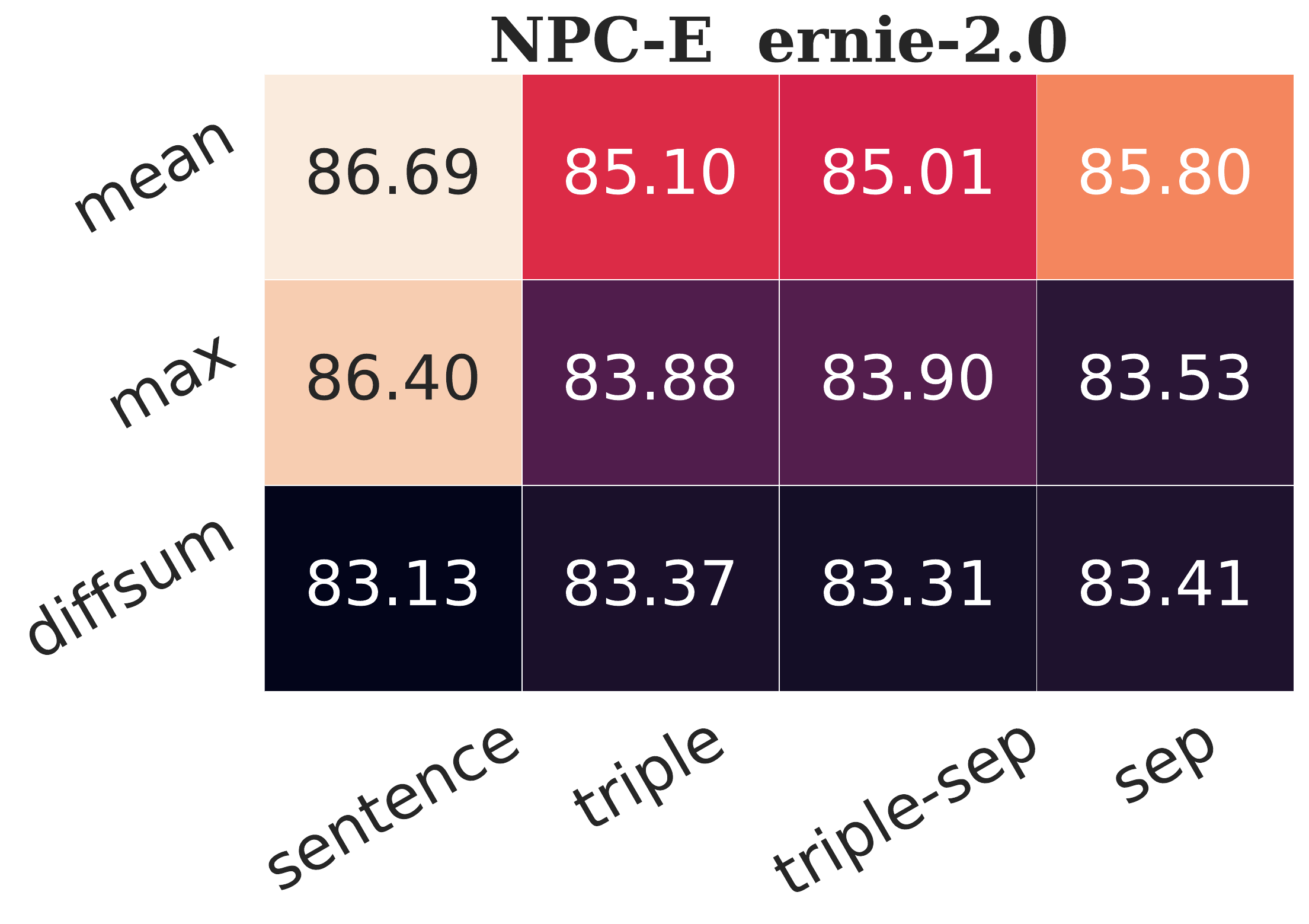}
     \end{subfigure}
     \hfill
     \begin{subfigure}[b]{0.32\textwidth}
         \centering
         \includegraphics[width=\textwidth]{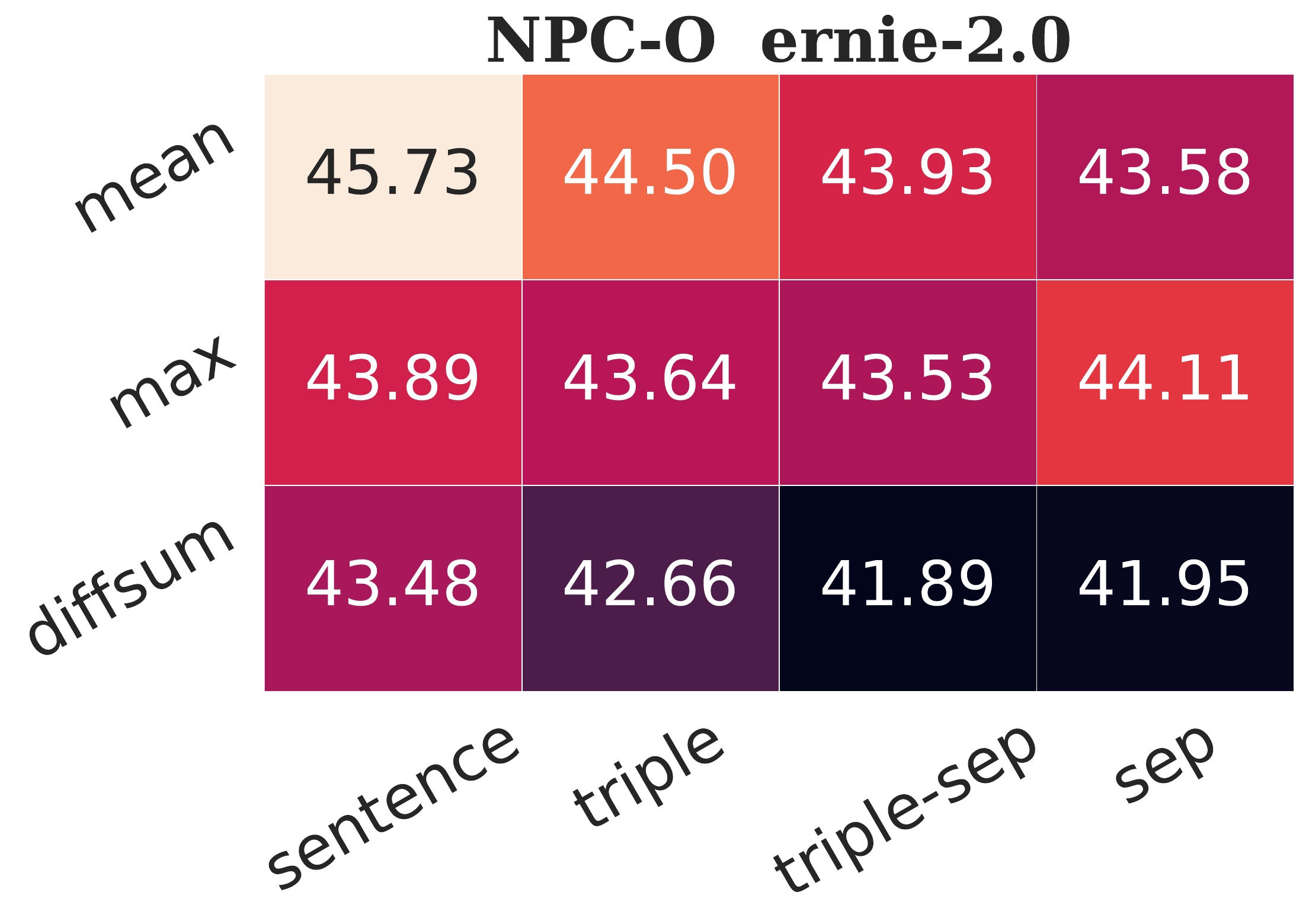}
     \end{subfigure}
     \hfill
     \begin{subfigure}[b]{0.32\textwidth}
         \centering
         \includegraphics[width=\textwidth]{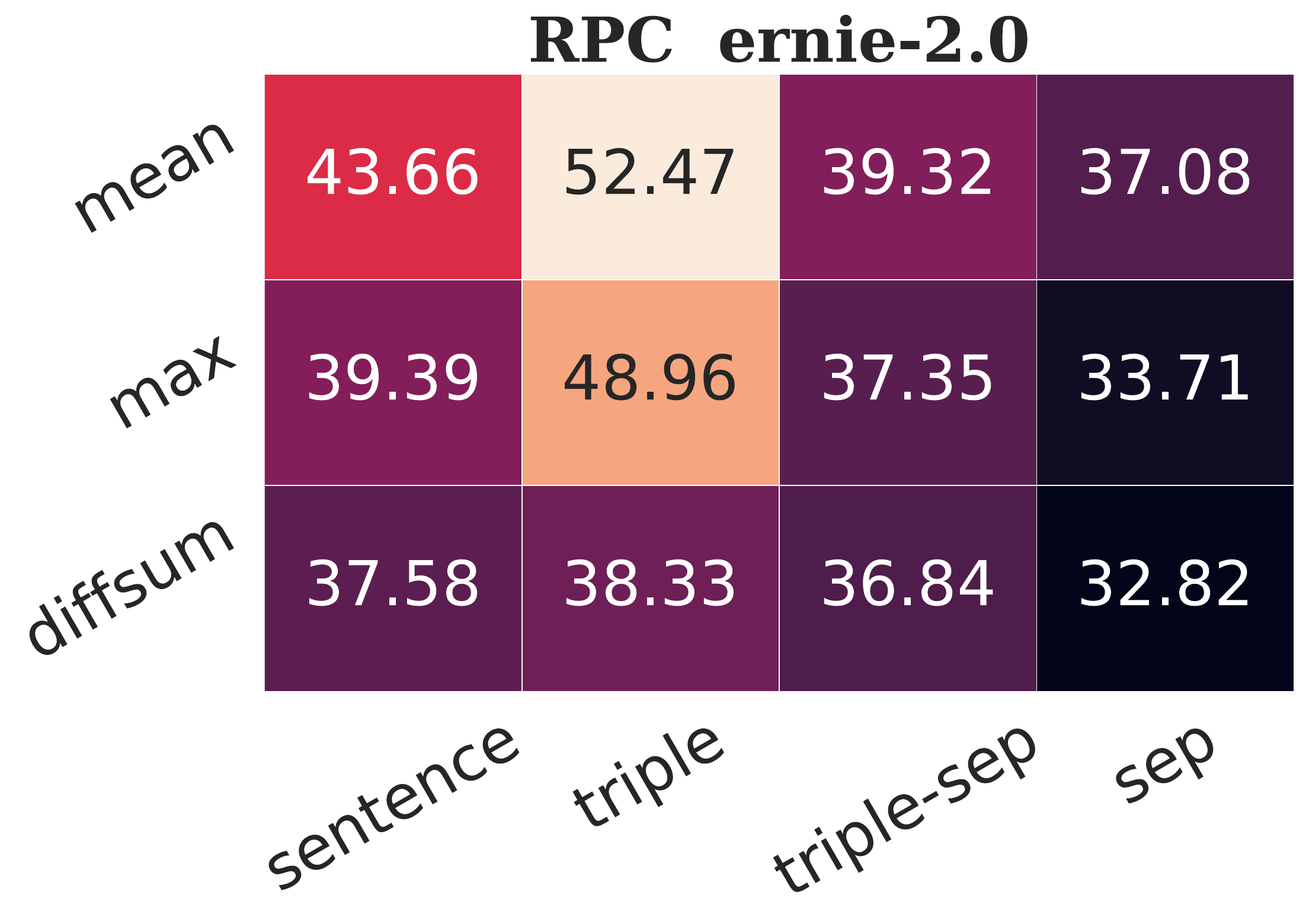}
     \end{subfigure}
        \caption{Comparison on different input forms and span representations for tasks and PLMs.}
        \label{fig:encode_all}
\end{figure*}

 We compare the performance of encoding strategies averaged on all subtask metrics and PLM models in Figure \ref{fig:res_encode}, and the task-specific and model-specific comparison of encoding strategies are shown in Figure \ref{fig:encode_all}. \textit{sentence} is the best input form in general, probably because it is easier for a PLM to encode a valid sentence and the source sentence contains more context. \textit{sep} is the worst on the RPC because it separately encodes the RPs and NPs. However, it is comparable to \textit{triple-sep} and \textit{triple} on NPC-E because NPC-E requires less context. \textit{mean} is the best strategy for phrase representation, which is consistent with the results obtained by \cite{toshniwal2020cross}\footnote{ \citet{toshniwal2020cross} shows that mean pooling is best for named entity labeling, which is a semantic task for NPs.}, and \textit{diffsum} is a bad choice for phrase canonicalization.

\section{Layerwise PLM Performance}
\label{app:layers}
\begin{figure}[h]
     \centering
     \begin{subfigure}[b]{0.45\textwidth}
         \centering
         \includegraphics[width=\textwidth]{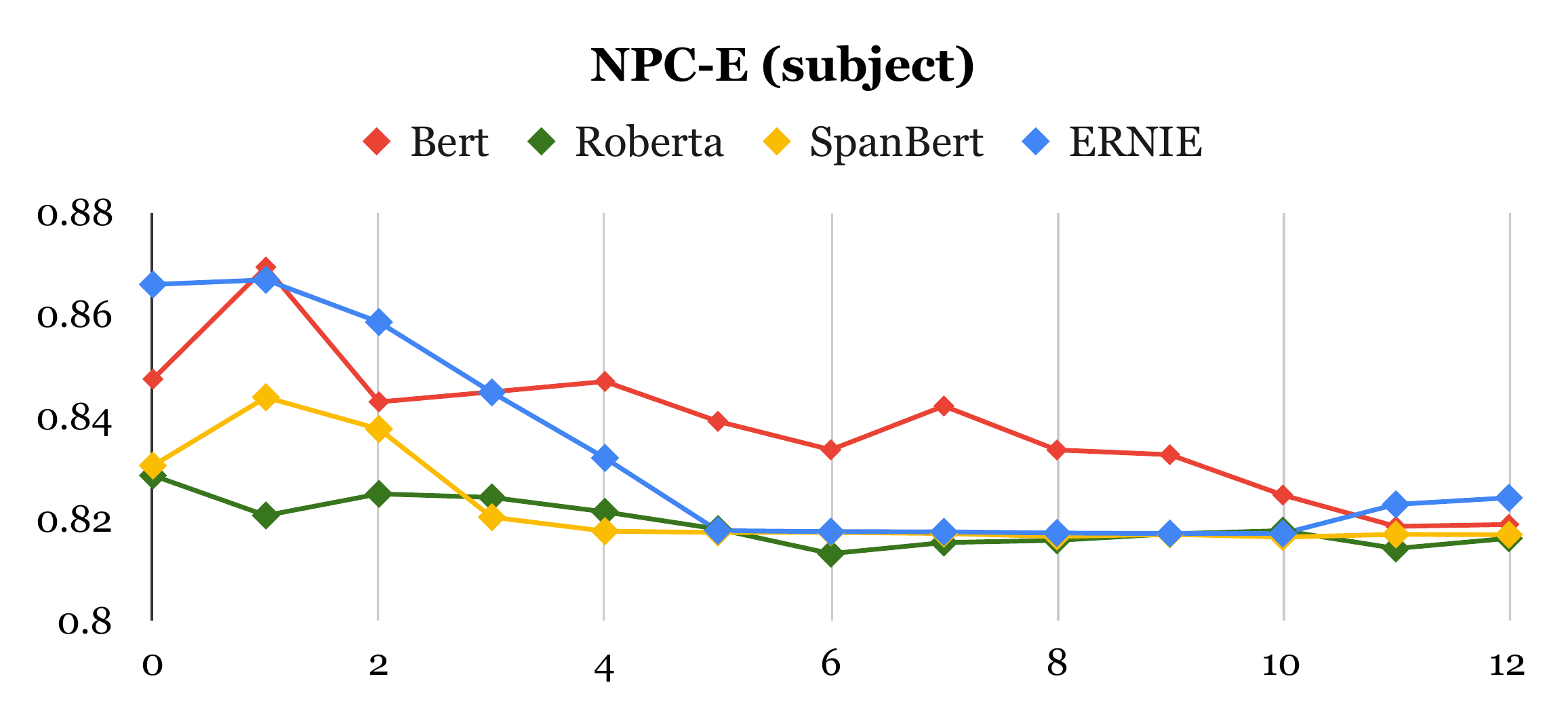}
     \end{subfigure}
     \hfill
     \begin{subfigure}[b]{0.45\textwidth}
         \centering
         \includegraphics[width=\textwidth]{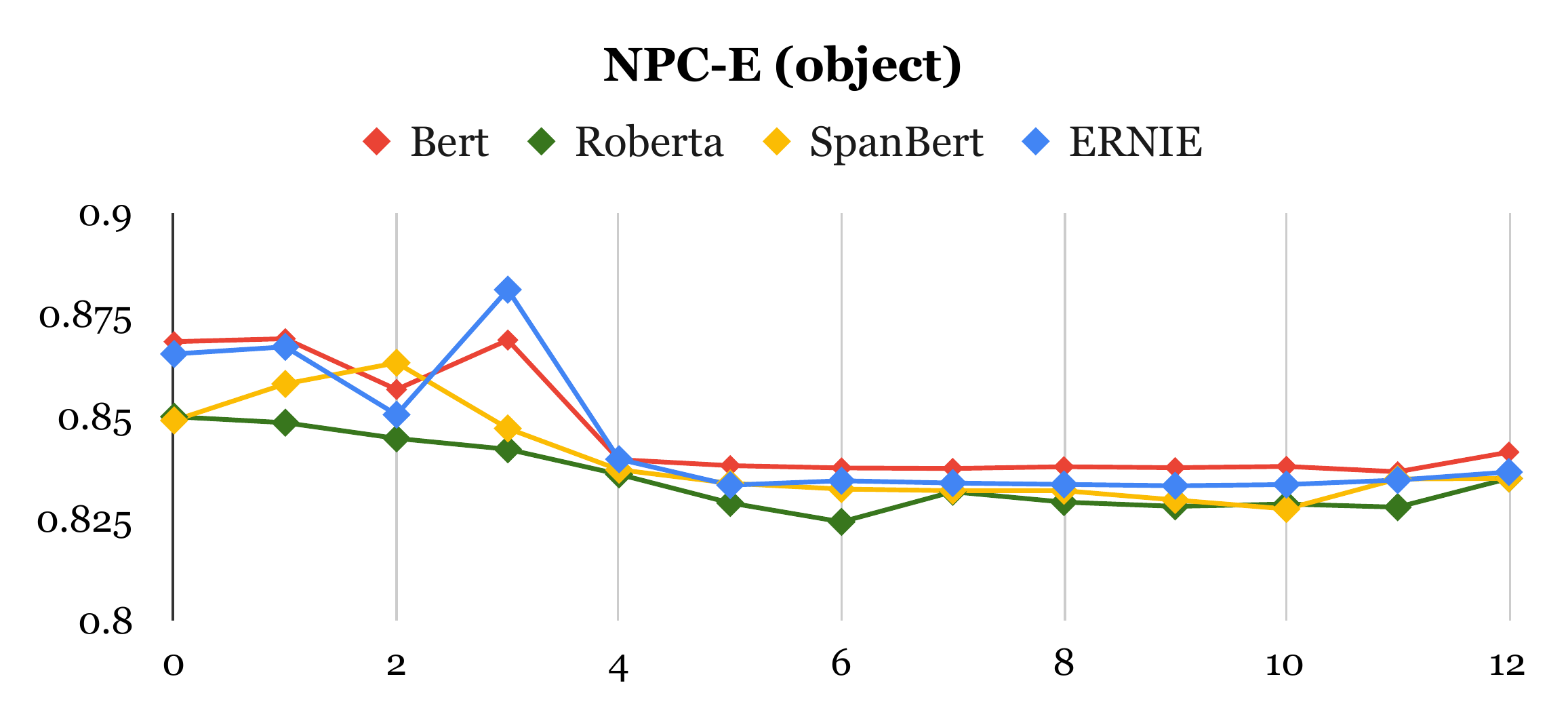}
     \end{subfigure}
     \hfill
     \begin{subfigure}[b]{0.45\textwidth}
         \centering
         \includegraphics[width=\textwidth]{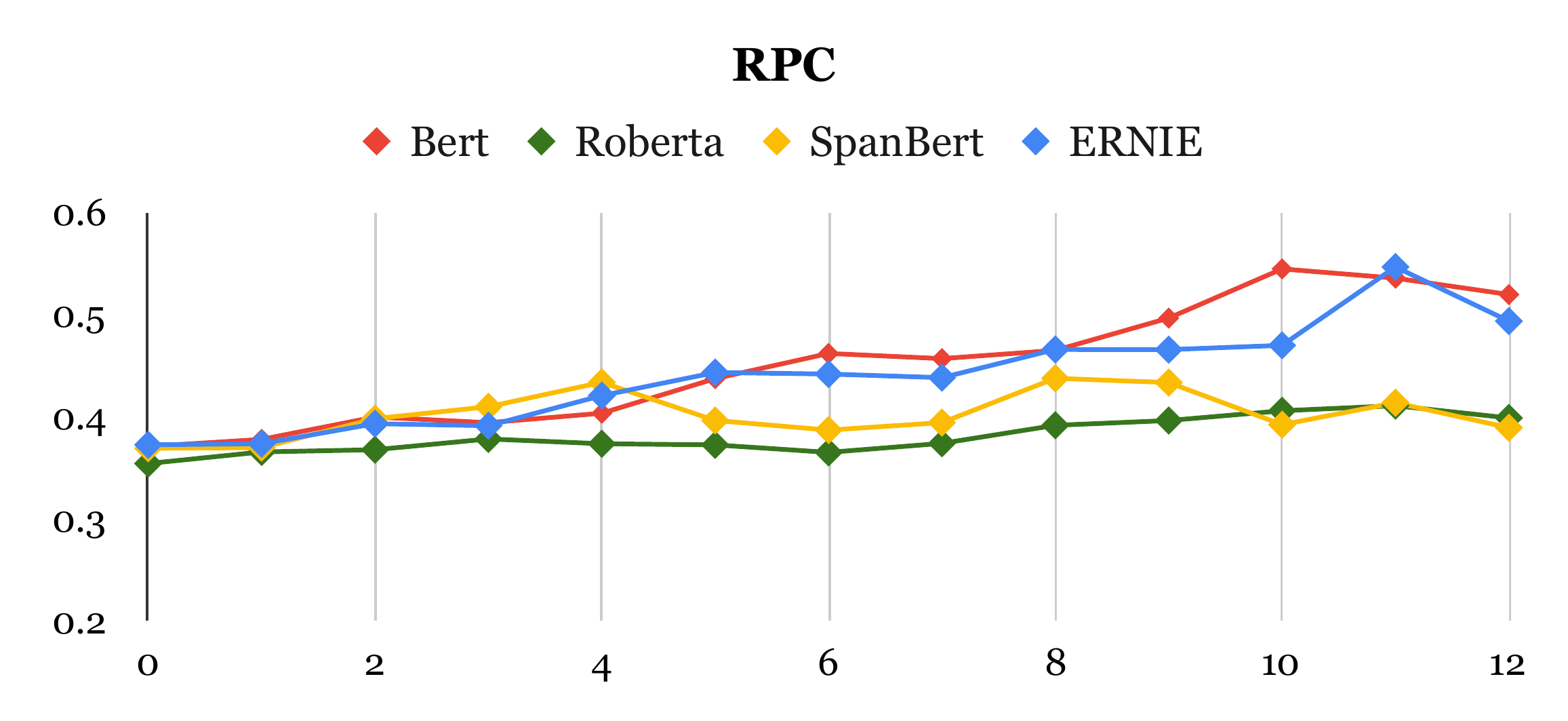}
     \end{subfigure}
     \hfill
     \begin{subfigure}[b]{0.45\textwidth}
         \centering
         \includegraphics[width=\textwidth]{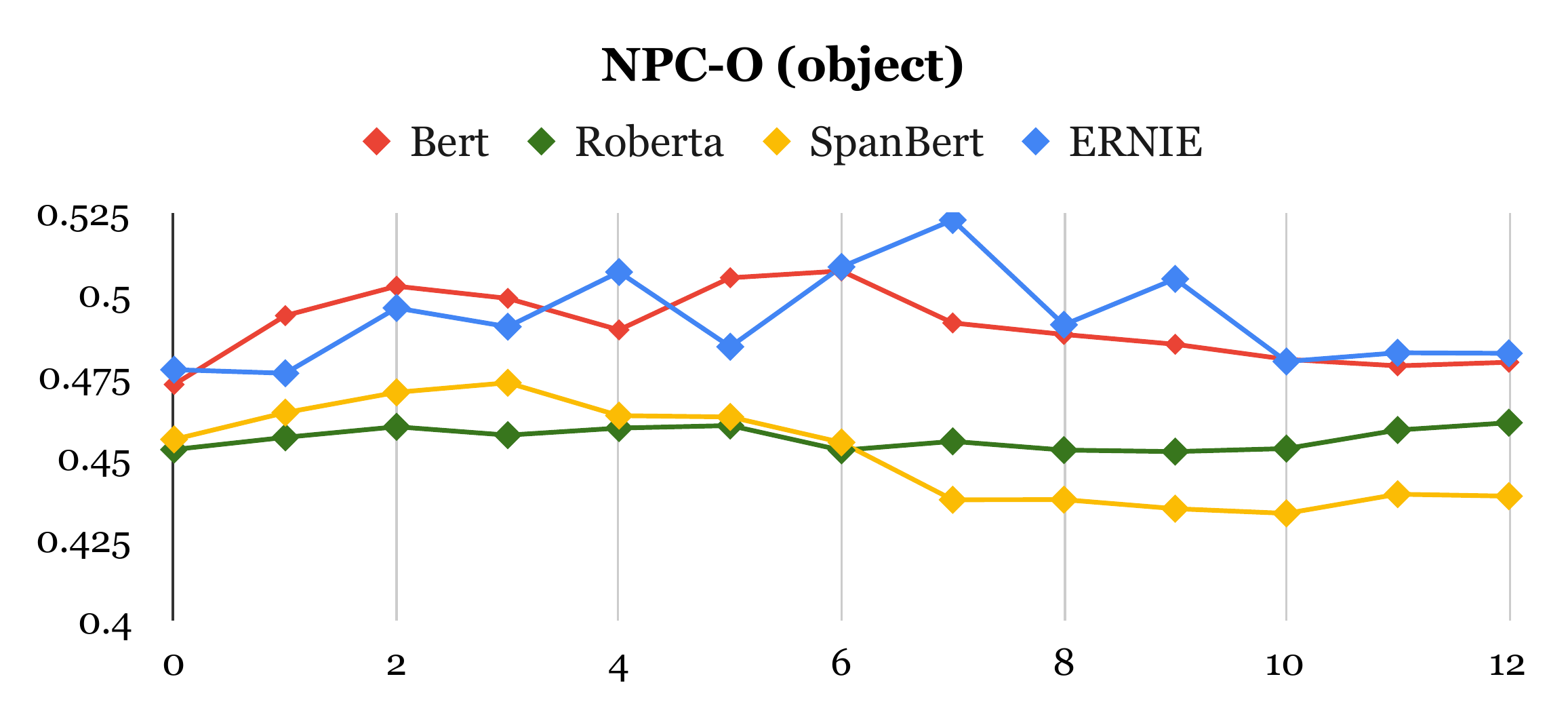}
     \end{subfigure}
     \hfill
     \begin{subfigure}[b]{0.45\textwidth}
         \centering
         \includegraphics[width=\textwidth]{img/layers/NPC-O-object.pdf}
     \end{subfigure}
     \hfill
     \begin{subfigure}[b]{0.45\textwidth}
         \centering
         \includegraphics[width=\textwidth]{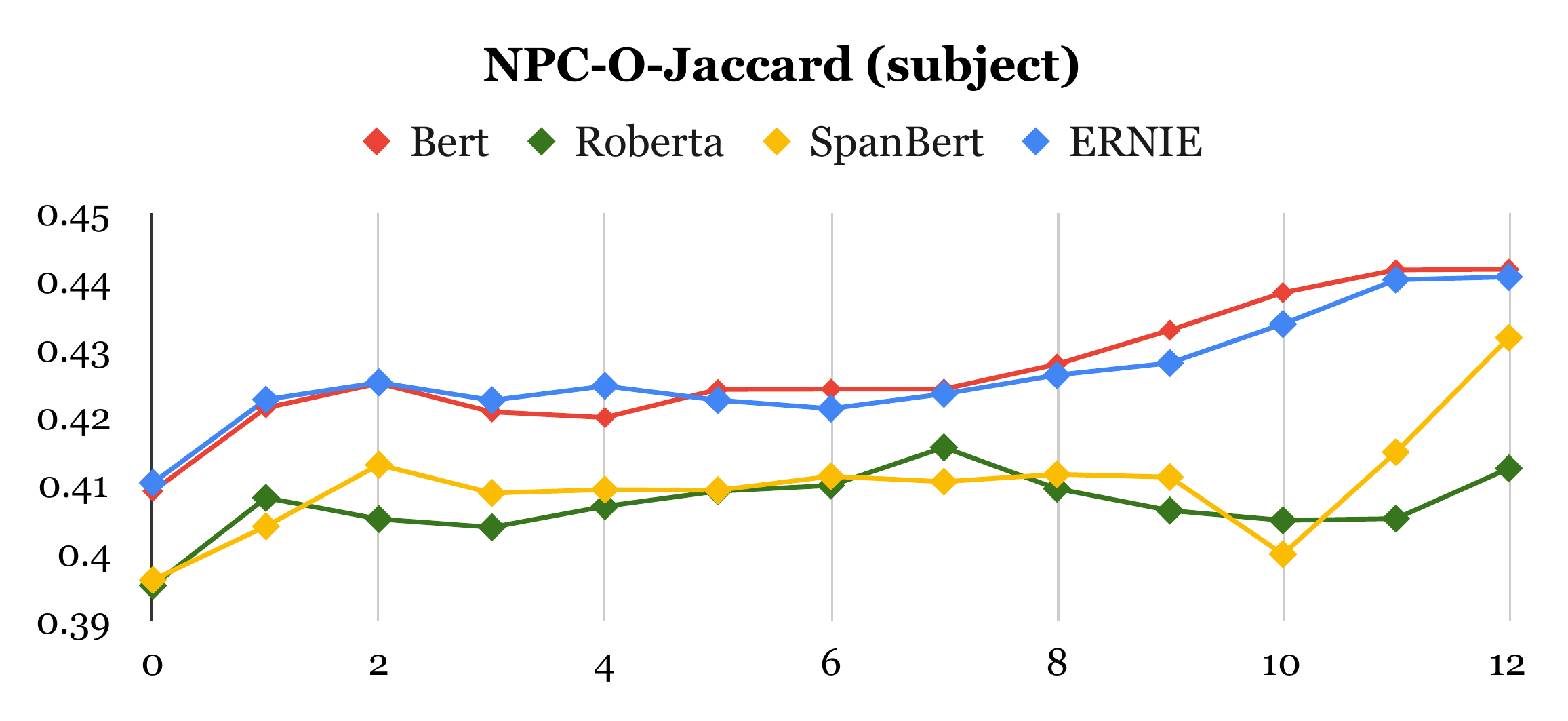}
     \end{subfigure}
     \hfill
     \begin{subfigure}[b]{0.45\textwidth}
         \centering
         \includegraphics[width=\textwidth]{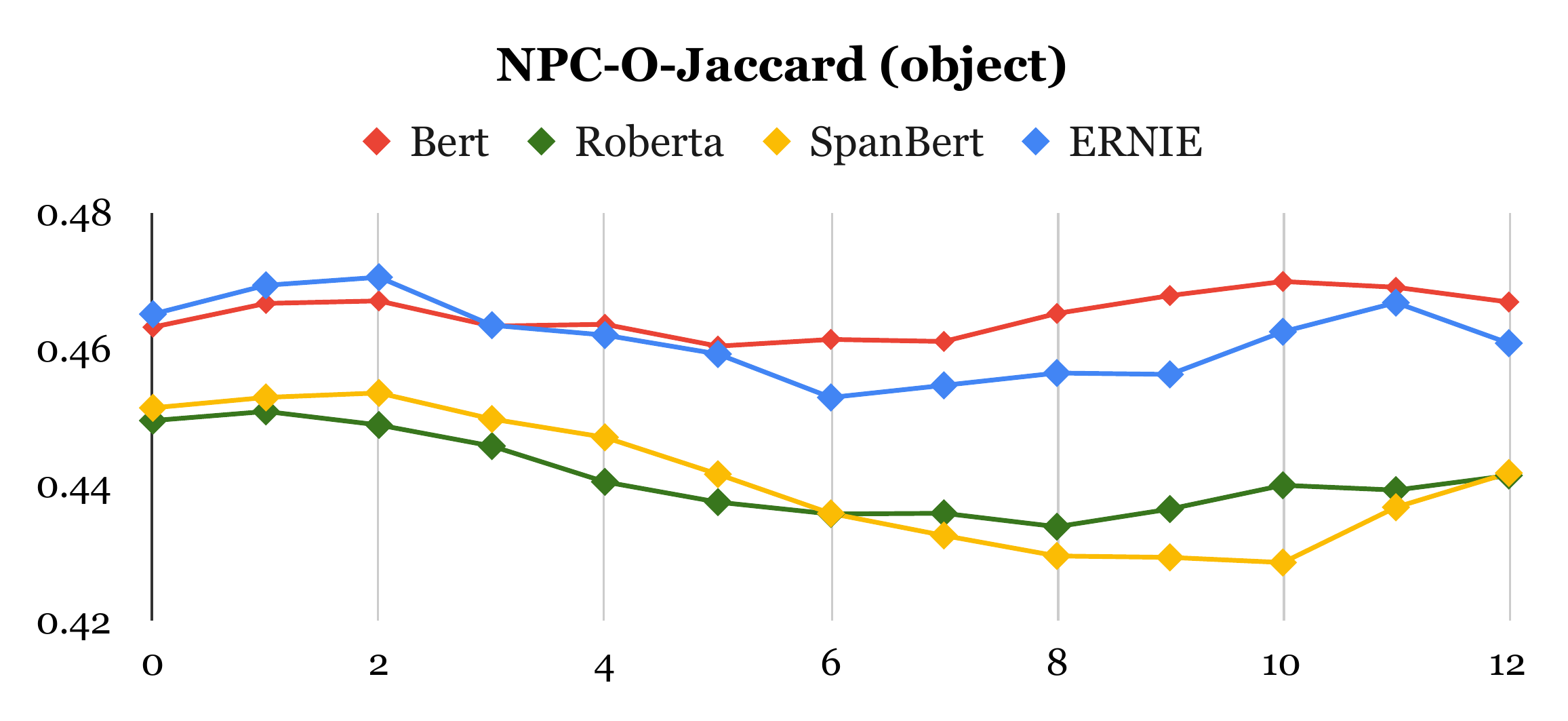}
     \end{subfigure}
        \caption{Layerwise performances}
        \label{fig:layers_all}
\end{figure}

We show the layerwise performance for all PLMs (base) on all subtasks in Figure \ref{fig:layers_all}, and find different layers perform differently on three subtasks. We empirically find that lower layers [1,2,3] perform well for NPC-E, upper layers [10,11,12] perform best for RPC and NPC-O (subj), while middle layers [3,4,5,6,7] perform relatively better on NPC-O (obj). As context-specificity increases in upper layers \cite{ethayarajh-2019-contextual}, these results make sense as NPC-E requires less context while RPC and NPC-O need more context.

\section{Triple Pretraining}
\label{app:triple}
\begin{figure}[!h]
    \centering
    \scalebox{0.28}{
    \includegraphics{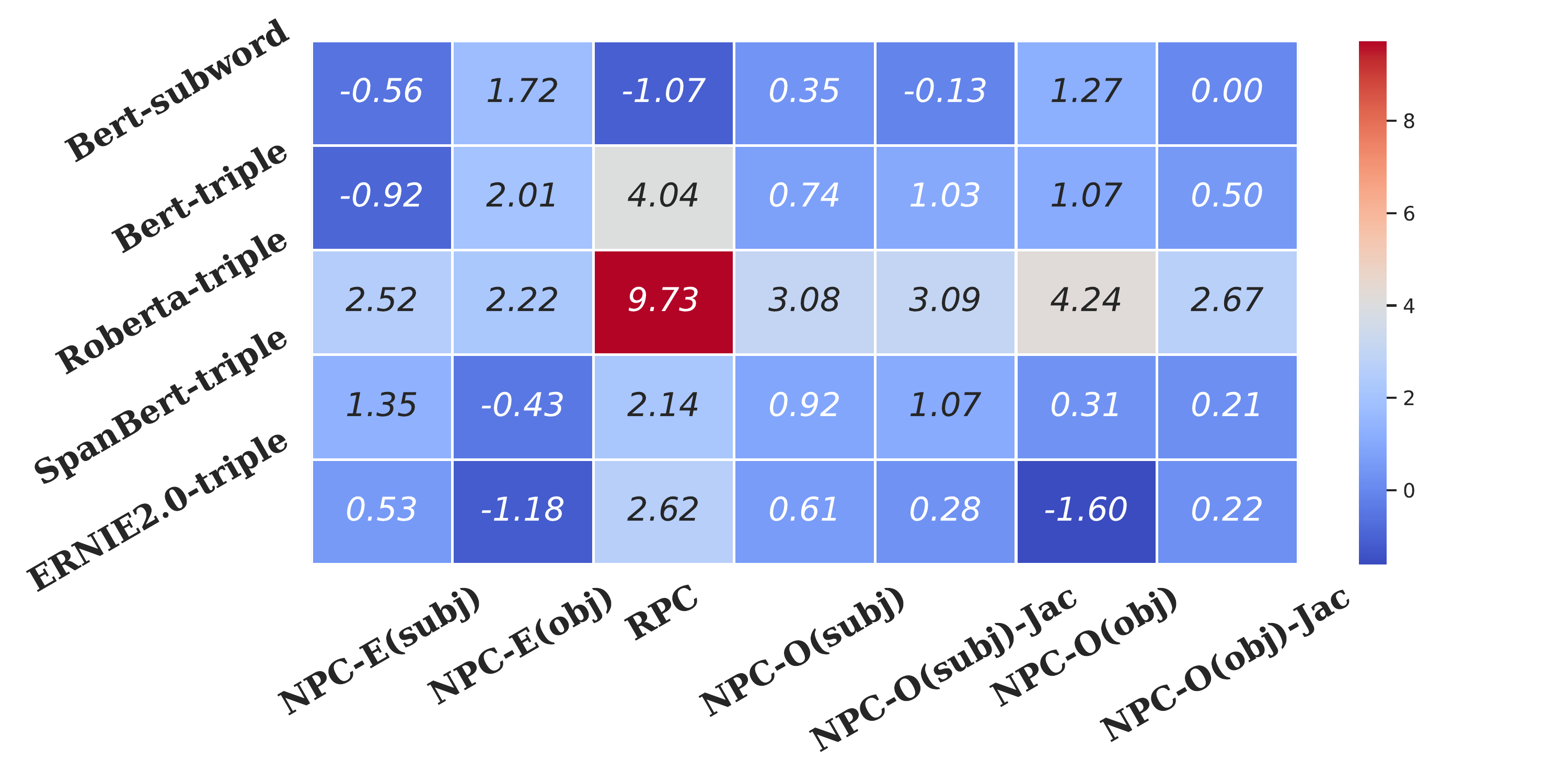}}
    \caption{Average performance difference after triple-level or causal subword-level pretraining for different PLMs and subtasks.}
    \label{fig:res_pretrain}
\end{figure}
We show the average performance difference after triple-level or causal subword-level pretraining in Figure \ref{fig:res_pretrain} for different PLMs and subtasks.

\end{document}